\documentclass[journal]{IEEEtran}

\ifCLASSOPTIONcompsoc
  \usepackage[nocompress]{cite}
\else
  \usepackage{cite}
\fi

\ifCLASSINFOpdf
  \usepackage[pdftex]{graphicx}
\else
  \usepackage[dvips]{graphicx}
\fi

\usepackage{amsmath,amssymb}
\usepackage{graphicx}
\usepackage{float}
\usepackage{color}

\usepackage[ruled,vlined]{algorithm2e}
\usepackage{algorithmic}
\usepackage{multirow}
\usepackage{bbm}
\usepackage{hyperref}
\usepackage{subcaption}
\usepackage[section]{placeins}

\newcommand{\rulesep}{\unskip\ \vrule\ }

\graphicspath{{./Figures/}}
\DeclareMathOperator*{\argmax}{argmax}
\DeclareMathOperator*{\argmin}{argmin}
\DeclareMathOperator{\E}{\mathbb{E}}
\begin{document}

\title{Daedalus: Breaking Non-Maximum Suppression in Object Detection via Adversarial Examples}

\author{Derui Wang,
        Chaoran~Li,
        Sheng~Wen,
        Qing-Long Han, Fellow, IEEE,
        Surya~Nepal,
        Xiangyu Zhang,
        and~Yang~Xiang, Fellow, IEEE

\IEEEcompsocitemizethanks{\IEEEcompsocthanksitem {D. Wang, C. Li, S. Wen, Q.L. Han, and Y. Xiang} are with Swinburne University of Technology, Hawthorn, VIC 3122, Australia. D. Wang and C. Li are also with Data61, CSIRO, Australia (E-mail: \{deruiwang, chaoranli, swen, yxiang\}@swin.edu.au).
\IEEEcompsocthanksitem S. Nepal is with Data61, CSIRO, Epping, NSW 1710, Australia (E-mail: surya.nepal@data61.csiro.au).
\IEEEcompsocthanksitem Xiangyu Zhang is with Department of Computer Science, Purdue University, West Lafayette, IN 47907, United States (E-mail: xyzhang@cs.purdue.edu).
}}

\markboth{XXX}%
{Shell \MakeLowercase{\textit{et al.}}: Bare Demo of IEEEtran.cls for Computer Society Journals}

\IEEEtitleabstractindextext{%
\begin{abstract}
This paper demonstrates that Non-Maximum Suppression (NMS), which is commonly used in Object Detection (OD) tasks to filter redundant detection results, is no longer secure. Considering that NMS has been an integral part of OD systems, thwarting the functionality of NMS can result in unexpected or even lethal consequences for such systems. In this paper, an adversarial example attack which triggers malfunctioning of NMS in end-to-end OD models is proposed. The attack, namely \texttt{Daedalus}, compresses the dimensions of detection boxes to evade NMS. As a result, the final detection output contains extremely dense false positives. This can be fatal for many OD applications such as autonomous vehicles and surveillance systems. The attack can be generalised to different end-to-end OD models, such that the attack cripples various OD applications. Furthermore, a way to craft robust adversarial examples is developed by using an ensemble of popular detection models as the substitutes. Considering the pervasive nature of model reusing in real-world OD scenarios, Daedalus examples crafted based on an \textit{ensemble of substitutes} can launch attacks without knowing the parameters of the victim models. Experimental results demonstrate that the attack effectively stops NMS from filtering redundant bounding boxes. As the evaluation results suggest, Daedalus increases the false positive rate in detection results to $99.9\%$ and reduces the mean average precision scores to $0$, while maintaining a low cost of distortion on the original inputs. It is also demonstrated that the attack can be practically launched against real-world OD systems via printed posters.
\end{abstract}

}

\maketitle

\IEEEdisplaynontitleabstractindextext

%
\IEEEpeerreviewmaketitle

\section{Introduction}
\begin{figure}[t!]
\center
\includegraphics[width=0.9\linewidth]{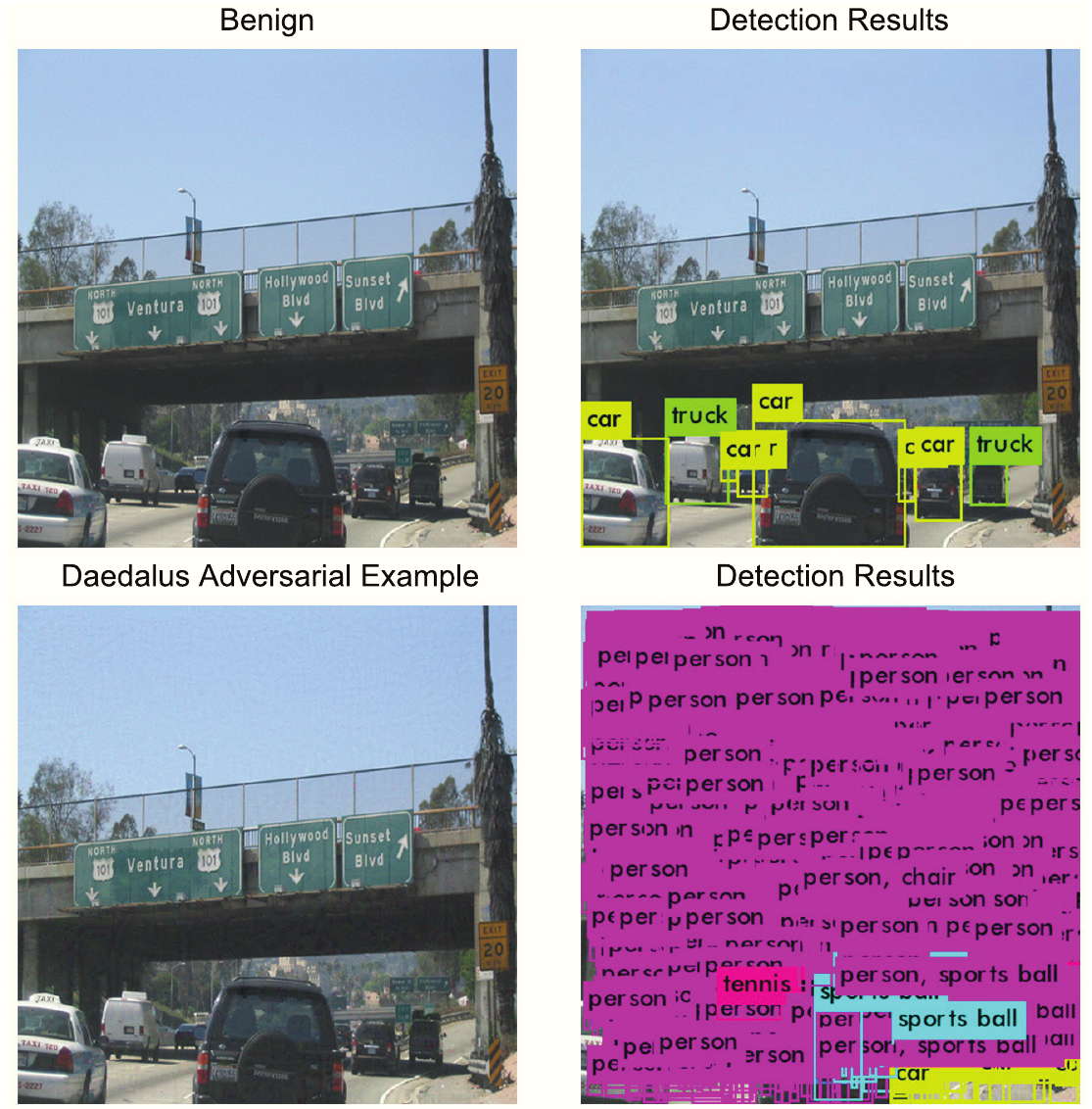}
\caption{A demo of the digital Daedalus attack on YOLO-v3. The upper-left image is an original image taken from the COCO dataset~\cite{lin2014microsoft}. The OD results are shown in the upper-right image. An adversarial example as well as the OD results of the adversarial example are shown in the second row. The adversarial example significantly increases the number of final detection boxes after NMS and makes the detector non-viable. However, the perturbation on the adversarial example is still imperceptible.}
\label{Examples}
\end{figure}

\IEEEPARstart{O}{bject} Detection (OD) is a fundamental operation in the field of computer vision and is repeatedly used in the areas of autonomous vehicles, robotics, surveillance system, and biometric authentications. Convolutional Neural Network (CNN) is embedded in the core of many state-of-the-art OD models. These models can be classified into two broad categories: single-stage detection models (\textit{e.g.}, SSD~\cite{liu2016ssd}, RetinaNet~\cite{lin2017focal}, and You-Only-Look-Once (YOLO)~\cite{redmon2016you}) and two-stage detection models (\textit{e.g.}, R-CNN~\cite{girshick2014rich} and Fast R-CNN~\cite{girshick2015fast}). Single-stage ones become successful since they perform end-to-end detection which has low computational overhead and is competitively accurate. Current end-to-end detection models are mostly built upon Fully Convolutional Network (FCN) backbones. However, FCNs output a large number of redundant detection results that should be filtered before the final results are produced.

To filter the redundant detection boxes, Non-Maximum Suppression (NMS) is performed as a key step in the OD procedure~\cite{neubeck2006efficient,cong2018speedup}. NMS has been an integral component in many OD algorithms for around 50 years. It relies on non-differentiable, hard-coded rules to discard redundant boxes. Beyond the OD task, NMS and its variants are also being used in applications such as face recognition and keypoint detection~\cite{Schroff2015FaceNet,wu2018face}. Some recent works tried to use a differentiable model to learn NMS~\cite{hosang2017learning}. Another work suggested using soft-NMS to improve the filtering performance~\cite{bodla2017soft}. Therefore, NMS is very important to OD systems, once the NMS component in an OD system does not work properly, the system fails.

CNN is known to be vulnerable to adversarial example attacks~\cite{Akhtar2018Threat}. For a human eye, an adversarial image can be indistinguishable from the original one, but it can cause misclassification in a CNN based classifier. Recently, adversarial examples have been crafted for OD and segmentation tasks~\cite{lu2017adversarial,xie2017adversarial,song2018physical,chen2018shapeshifter,thys2019fooling,zhao2019seeing}. Such attacks result in either the misclassification or the appearance/disappearance of the objects. On the other hand, false positive rate is a critical metric in OD missions such as inventory management~\cite{verma2016object} and military object recognition~\cite{yang2018deep}. An attack that boosts the false positive rate could be very lethal to these OD systems.

In this paper, we propose a novel adversarial example attack for NMS, which can be universally applied to any other end-to-end differentiable OD model. A demonstration of the attack is shown in Fig.~\ref{Examples}. As it can be seen, a properly crafted adversarial example can trigger malfunctioning of the NMS component in YOLO-v3, in a way that it outputs dense and noisy detection results. We name this attack \texttt{Daedalus} since it creates a maze composed of false positive detection results.

First, Daedalus is evaluated and analysed on how it can reduce the OD performance by breaking NMS\footnote{In this paper, the term \textit{breaking NMS} means that NMS cannot filter redundant bounding boxes detected from an adversarial example.}. Furthermore, Daedalus is adapted for launching black-box attack. Based on a census in Section~\ref{census:sec}, we observe two facts: 1) current OD tasks are thriving on a few types of OD models, even many models have been proposed; 2) current OD tasks often load trained parameters into a model instead of reparameterising the model. Therefore, if an attacker can generate a universal adversarial example which works effectively against all the popular models, the example has a high chance to successfully sabotage a black-box model. Herein, we propose using an \textit{ensemble of substitutes} (\textit{i.e.}, an ensemble of popular detection models as the substitutes) in the Daedalus attack to break the black-box OD models. We exploit the latest version of YOLO (\textit{i.e.}, YOLO-v3~\cite{redmon2018yolov3}), SSD~\cite{liu2016ssd}, and RetinaNet~\cite{lin2017focal} as the victims. As suggested by the experiments, universal Daedalus examples generated by using an ensemble of popular OD models can launch attacks in a black-box scenario. To further enhance the viability of the attack, we instantiate the attack as a physical poster that can break NMS of OD models in the real world. In summary, the contributions of this paper are listed as follows: 

\begin{itemize}
\item\textit{A novel adversarial example attack which aims at breaking NMS (unlike the existing state-of-the-art attacks that merely target misclassification) is proposed. As a consequence, the attack creates a large number of false positives in the detection results;}
\item\textit{An ensemble-of-substitutes attack is proposed to generate universal adversarial examples which can make NMS malfunction in various models. In such way, black-box attacks can be partially converted to white-box attacks;}
\item\textit{Physical posters containing Daedalus perturbations are crafted to practically attack real-world OD applications.}
\end{itemize}

The paper is organised as follows: Section~\ref{Primer} introduces the background knowledge of adversarial example, object detection, and NMS. Section~\ref{Method} describes the design of Daedalus. Section~\ref{Evaluation} compares and selects loss functions from the candidates, and systematically evaluates the performance of the Daedalus attack. Section~\ref{Discussion} discusses the possibility of applying Daedalus adversarial examples on soft-NMS and a group of black-box OD models. Moreover, we propose using a poster of the Daedalus perturbation to attack real-world OD applications. We also post the dilemma of defending against Daedalus in this section. Section~\ref{Related_work} reviews the related work relevant to this study. Section~\ref{Conclusion} draws a conclusion on the paper and presents our future work.

\section{Preliminaries}\label{Primer}
\subsection{Adversarial examples}
Given a DNN model $F$ and a benign data example $x$, an attacker can craft an adversarial example of $x$ by altering the features of $x$ by adding a perturbation $\delta$, such that $F(x+\delta)\neq F(x)$. Generating adversarial examples can be interpreted as optimising the example features towards adversarial objectives. Gradient-based optimisation methods are commonly employed to search for adversarial examples. The optimisation objective can be generally parsed as follows:

\begin{align}
&\argmin_{\delta}{|\delta|_{p}+c\cdot L(F(x+\delta), y)}\\
&\ s.t.\ \ x+\delta\in [0,1]^{n}, \nonumber
\end{align}
wherein $L$ is an adversarial objective set by the attacker, and $y$ is either a ground truth output or a desired adversarial output for $F$. $x+\delta$ produces an adversarial example $x_{adv}$. $|\delta|_{p}$ is the allowed distortion bounded by the $p$-norm of $\delta$. A constant $c$ is adopted to balance the adversarial loss and the distortion. This box-constrained non-convex optimisation problem is typically solved by gradient descent methods.

\subsection{A census of object detection models}\label{census:sec}
To better understand the impact of attacks against OD tasks, we first need to overview the distribution of models used in OD. We conducted a census of OD projects on Github. We crawled project names, project descriptions, and readme files from $1,696$ OD-related repositories on Github. Next, based on the reviewed models in \cite{liu2018deep}, $45$ released OD models were selected as keywords. For each repository, we counted the keywords in the readme file, and selected the most frequent one as the model adopted for the repository. Finally, we grouped the repositories by models. The ranking of models that appeared more than $50$ times is presented in Table~\ref{census}. Based on the analysis, $1,561$ out of $1,696$ repositories use models from Table~\ref{census}. The observation suggests that models and backbones used for OD tasks are highly limited in number. Moreover, we analysed the source codes from the repositories to search for operations of loading trained models and parameters. We observed that almost all the source codes provide an interface for loading pre-trained parameters into the model backbones. We further investigated names of the pre-trained models and parameters. We found that many repositories use the same pre-trained parameters in their models.

\begin{table}[t!]
\caption{Object detection Models on Github}
\label{census}
\centering
\resizebox{0.7\columnwidth}{!}{%
\begin{tabular}{c c c c c}
\hline
Rank & Model & Popularity\\
\hline
1 & SSD & 605\\
2 & YOLO-v3 & 251\\
3 & Faster R-CNN & 236\\
4 & R-CNN & 178\\
5 & YOLO-v2 & 164\\
6 & RetinaNet & 75\\
7 & YOLO-v1 & 52\\
\hline
Popular models/Total & & 1,561/1,696\\
\hline
\end{tabular}
}
\end{table}

Based on the observations, black-box attacks towards OD models can be partially converted to white-box attacks. An attacker can first craft a universal adversarial example which breaks all the popular white-box OD models, and then launch attacks on black-box models using the example. The attack has a high potential to break the black-box models since that they may have the same parameters with the white-box models.

\subsection{Object detection}\label{YOLOv3}
Given an input frame, an end-to-end OD model outputs detected regression boxes and classification probabilities through a FCN. Herein, we adopt the latest version of You Only Look Once (YOLO-v3) detector~\cite{redmon2016you} as an example to introduce the knowledge of end-to-end detection algorithms. YOLO has achieved the state-of-the-art performance on standard detection tasks such as PASCAL VOC~\cite{everingham2010pascal} and COCO~\cite{lin2014microsoft}. YOLO has been studied extensively in the research of video OD, robotics, and autonomous cars~\cite{shafiee2017fast,lu2017efficient, zhang2017real}.

The backbone of YOLO-v3 is a 53-layer FCN (\textit{i.e.}, Darknet-53). YOLO-v3 segments the input image using three mesh grids in different scales (\textit{i.e.}, $13\times 13$, $26\times 26$, $52\times 52$). The model outputs detection results on these three scales. Each grid contains three anchor boxes, whose dimensions are obtained by K-means clustering on the dimensions of ground truth bounding boxes. Each grid outputs three bounding box predictions. Henceforth, there are totally $10647$ bounding boxes in the output of YOLO-v3. The Darknet-53 backbone outputs a feature map which contains bounding box parameters, the box confidence, and object class probabilities. Specifically, for each bounding box, the feature map includes its height $t_h$, width $t_w$, center coordinates $(t_x$, $t_y)$, as well as the class probabilities $p_1, p_2,..., p_n$ and the objectness $t_0$. $t_0$ is the confidence that a box contains an object. Given the feature map, the position of each detection box is then calculated based on the anchor box dimension priors $p_w$ and $p_h$, and centre offsets $(c_x, c_y)$ from the top-left corner of the image. The final box dimension and position will then be:

\begin{align}
b_x &= c_x + \theta{(t_x)} \nonumber \\
b_y &= c_y + \theta{(t_y)} \\
b_w &= p_w{e^{t_w}} \nonumber \\
b_h &= p_h{e^{t_h}} \nonumber
\end{align}

where $b_w$ is the box width, and $b_h$ is the box height. $b_x$ and $b_y$ are the coordinates of the box center. The box confidence $b_0$ is calculated as:

\begin{equation}
b_0 = t_0\cdot \max\{p_1, p_2,..., p_n\}
\end{equation}

YOLO-v3 finally outputs the bounding boxes, the box confidences, and the class probabilities. The boxes whose objectness is below a preset threshold will be discarded. The outputs will then be processed by NMS to generate the final detection results.

\subsection{Non-Maximum Suppression (NMS)}

NMS is integrated into OD algorithms to filter detection boxes. NMS makes selections based on the Intersection over Union (IoU) among detection boxes. IoU measures the ratio of the overlapped area over the union area between two boxes. NMS works in two steps: 1) For a given object category, all of the detected bounding boxes in this category (\textit{i.e.}, candidate set) are sorted based on their box confidence scores from high to low; 2) NMS selects the box which has the highest confidence score as the detection result, and then it discards other candidate boxes whose IoU value with the selected box is beyond the threshold. Then, within the remaining boxes, NMS repeats the above two steps until there is no remaining box in the candidate set. Suppose the initial detection boxes of category $c$ are $B^c={b^c_1, b^c_2, ..., b^c_n}$, the corresponding box confidence scores are $S^c={s^c_1, s^c_2, ..., s^c_n}$. Given an NMS threshold $N_t$, we can write the NMS algorithm as Algorithm~\ref{nms}:

\begin{algorithm}[h!]
\footnotesize
\caption{Non-Maximum Suppression}
\label{nms}
\textbf{Input}: $B^c$, $S^c$, $N_t$ \\
\textbf{Initialisation}: \\
$D^c \leftarrow \{\}$ \\
\While{$B^c \neq empty$}
{
	$m \leftarrow \argmax S^c$ \\
	$M \leftarrow b^c_m$\\
	$D^c \leftarrow D^c \cup M$ \\
	$B^c \leftarrow B^c-M$ \\
	\For{$b^c_i \in B^c$}
	{
		\If{$IoU(M, b^c_i) \geq N_t$}
		{
			$B^c \leftarrow B^c-b^c_i$ \\
			$S^c \leftarrow S^c-s^c_i$ \\
		}
	}
}
\textbf{Output}: $D^c$, $S^c$.
\end{algorithm}

NMS recursively discards the redundant detection boxes from the raw detection proposals in all object categories, until all raw proposals have been processed.

\section{The Daedalus attack}\label{Method}
The proposed Daedalus attack is introduced in this section. To clarify the intuition behind the attack, we first analyse the vulnerability that we exploit to break NMS. Then, we introduce the general optimisation techniques used for making adversarial examples. Subsequently, we propose a series of adversarial loss functions to generate Daedalus adversarial examples. At last, we assemble the algorithms together to present the Daedalus attack, which can force a detection network to output an adversarial feature map that causes the malfunctioning of NMS.

\subsection{Attacking NMS}\label{AttackNMS}
We first introduce the NMS vulnerability which leads to the attack. NMS filters boxes based on their IoUs. Specifically, after the raw detection boxes are obtained, the boxes with a confidence score below a given threshold will be discarded. Next, for a given object class and a box of this class with the highest box confidence score (\textit{i.e.}, the final detection box), NMS will discard other remaining boxes based on their IoUs with the final detection box. This mechanism makes NMS vulnerable to attacks that suppress the IoUs among output bounding boxes. Once the IoUs are suppressed below the required threshold for NMS filtering, NMS can no longer function in the regular way. In this case, most of the redundant bounding boxes will be kept in the final detection results. 

\begin{figure}[t!]
\center
\includegraphics[width=0.9\linewidth]{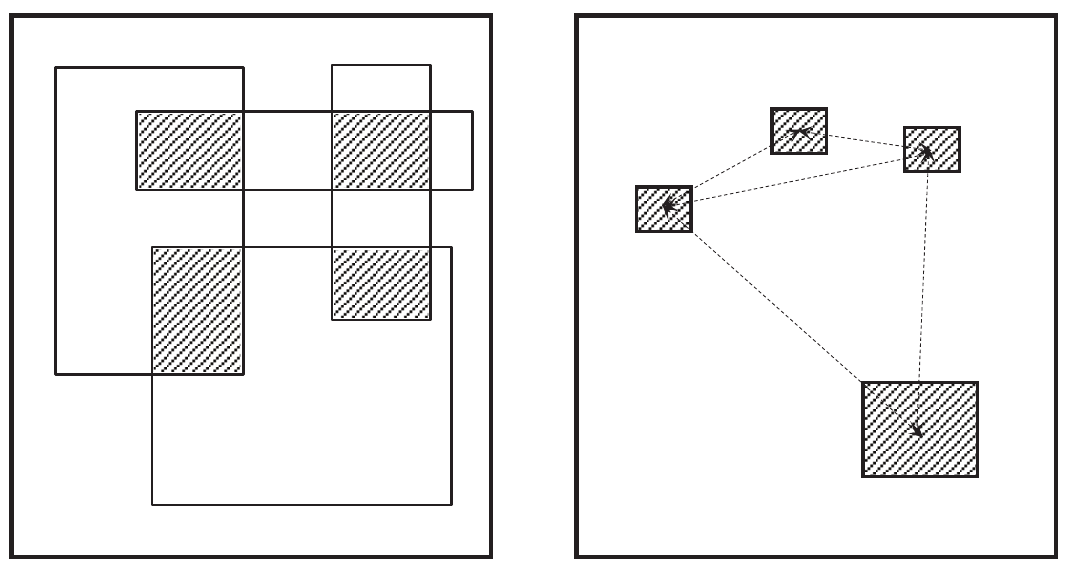}
\caption{The schemes we used to attack NMS. The scheme on the left side directly minimises IoUs between all box pairs. The second one minimises the dimension for all boxes, and maximises the Euclidean distance among box centres for all boxes. The third scheme takes a low-cost approximation, which only minimises the dimension for all boxes.}
\label{attack_NMS}
\end{figure}

Three elemental schemes are considered to break NMS. First, for algorithms such as YOLO-v3, it is necessary to make most of the bounding boxes survive the first round of filtering that discards boxes based on the box confidences. Hence, we need to maximise the box confidences. Second, to compress the IoUs, we can directly minimise the IoU for each pair of boxes. Alternatively, we can minimise the expectation of the box dimension over all the bounding boxes, and maximise the expectation of the Euclidean distance between box centres over all pairs of boxes. The schemes are illustrated in Fig.~\ref{attack_NMS}.

\subsection{Generating adversarial examples}
A benign example is denoted as $x$. An adversarial example of $x$ is denoted as $x'$. The adversarial perturbation is $\delta=x'-x$. Therefore, the distortion calculated based on $p$-norm of the perturbation is $\|\delta\|_{p}$. We can then formulate the attack as the following optimisation problem:

\begin{align}\label{obj_func}
&\argmin_{\delta}{\|\delta\|_{p}+c\cdot f(x+\delta)}\\
&s.t.\ \ x+\delta\in [0,1]^{n}, \nonumber
\end{align}
wherein $f$ is an adversarial loss function. We will propose a set of potential loss functions $f$ in the next section. $c$ is a constant to balance the distortion and the adversarial loss $f$. $\delta$ is the adversarial perturbation. An adversary optimises $\delta$ to search for an adversarial example whose pixel values are bounded between 0 and 1. This is a typical box-constraint optimisation problem. To bound the pixel values of the generated adversarial examples between 0 and 1, we adopt the idea of changing variables from the paper \cite{carlini2017}. We change the optimisation variables from $\delta$ to $\omega$. There are different functions we can use for changing the variables (\textit{e.g.}, sigmoid function, arctan function, hyperbolic tangent function, \textit{etc.}) Among these functions, the hyperbolic tangent function generally produces higher gradients than others. Henceforth, to make the optimisation converge faster, we select the hyperbolic tangent function to change the variables. For $\delta_i$ in $\delta$, we can apply a hyperbolic tangent function such that:

\begin{equation}\label{tanh}
\delta_i = \frac{1}{2}(tanh(\omega_i)+1) - x_i.
\end{equation}

We then optimise the above task based on the new variables $\omega$. We apply binary search to find the best $c$ during the optimisation. Finally, the adversarial perturbation $\delta$ can be calculated easily through the inverse of Function~\ref{tanh}. 

\subsection{Adversarial loss functions}\label{losses}
In this section, three loss functions are designed to find adversarial examples which trigger malfunctioning of NMS. Based on the discussion in Section~\ref{AttackNMS}, we formulate three potential adversarial loss functions. We will evaluate each loss function in Section~\ref{Evaluation}. There are totally $n$ detection boxes (\textit{e.g.}, $n=10647$ in YOLO-v3). Supposing the object detector can be regarded as a function $F$. Given an input image $x$, the outputs are $F(x) = \{B^x, B^y, B^w, B^h, B^0, P\}$. Herein, $B^x = \{b^x_0, b^x_1,...,b^x_n\}$, $B^y =\{b^y_0, b^y_1,...,b^y_n\}$, $B^w=\{b^w_0, b^w_1,...,b^w_n\}$, and $B^h = \{b^h_0, b^h_1,...,b^h_n\}$. They are the dimensions and coordinates of the $n$ output bounding boxes. $B^0 = \{b^0_0, b^0_1,...,b^0_n\}$ are the objectness scores and $P = \{p_0, p_1,...,p_n\}$ are the class probabilities as introduced in Section~\ref{YOLOv3}.

The attack can specify an object classified as category $\lambda$ to attack. If we want to attack multiple object categories, we can include these categories in a set $\Lambda$. Based on the above discussion, we define three loss functions, $f_1$, $f_2$, and $f_3$, as follows:

\begin{align}\label{f1}
f_{1}= & \frac{1}{\|\Lambda\|}\sum_{\lambda\in \Lambda} \displaystyle \mathop{\E}_{i:argmax(p_i)=\lambda}\{{[b^0_i \cdot max(p_i)-1]^2}+\\ \nonumber
& \displaystyle \mathop{\E}_{j:argmax(p_j)=\lambda}{IoU_{ij}}\},
\end{align}
\begin{align}\label{f2}
f_{2}= & \frac{1}{\|\Lambda\|}\sum_{\lambda\in \Lambda} \displaystyle \mathop{\E}_{i:argmax(p_i)=\lambda}{\{[b^0_i \cdot max(p_i)-1]^2}+\\ \nonumber
& {(\frac{b^w_i\cdot b^h_i}{W\times H})^2 + \displaystyle \mathop{\E}_{j:argmax(p_j)=\lambda}\frac{1}{(b^x_i-b^x_j)^2+(b^y_i-b^y_j)^2}\}},
\end{align}
\begin{align}\label{f3}
f_{3}= & \frac{1}{\|\Lambda\|}\sum_{\lambda\in \Lambda} \displaystyle \mathop{\E}_{i:argmax(p_i)=\lambda}{\{[b^0_i \cdot max(p_i)-1]^2}+\\ \nonumber
& {(\frac{b^w_i\cdot b^h_i}{W\times H})^2\}},
\end{align}
wherein, $IoU_{ij}$ is the IoU between the $i$-th and the $j$-th bounding boxes. Box dimensions are scaled between $0$ and $1$, dividing $b^w$ and $b^h$ by the input image width $W$ and height $H$, such that the term is invariant towards changing input dimension.

In $f_1$, we first minimise the expectation of mean squared error between the box confidence and $1$ for all the boxes in which the detected objects are in the attacked category set $\Lambda$. Henceforth, the boxes will not be discarded due to low box confidence scores. In the second term, $f_1$ minimises the expectation of IoUs for all pairs of bounding boxes in $\Lambda$. When the IoUs fall below the NMS threshold, the boxes can evade the NMS filter. Alternatively, we minimise the expectation of box dimensions and maximise the expectation of box distances in $f_2$. $f_2$ approximates the effect of directly minimising IoU by compressing box dimensions and distributing boxes more evenly on the detected region.

Pairwise computation for obtaining IoUs and box distances could be expensive if the output feature map contains too many boxes. Therefore, we also propose $f_3$ as a more efficient loss function. $f_3$ only minimises the expectation of box dimensions instead of directly minimising the expectation of the IoUs. Minimising $f_3$ leads to smaller bounding boxes. When the boxes become small enough, there will be no intersection among the boxes. In other words, the IoUs among the boxes will be suppressed to zero. As a consequence, the boxes can avoid being filtered by NMS.

\subsection{Ensemble of substitutes}\label{ensemble_loss}
In this section, an ensemble method is developed to convert black-box attacks into white-box ones based on the observation on the current OD tasks (q.v. Section~\ref{census:sec}). The transferability of adversarial examples is usually considered as a pathway for launching black-box attacks. The transferability of misclassifying adversarial examples has been investigated in~\cite{gurbaxani2018traits}. However, instead of causing misclassification, the Daedalus attack has a different attacking purpose (\textit{i.e.}, disabling NMS). It is difficult to transfer Daedalus examples among OD models with different backbones since the feature maps extracted by the backbones are highly divergent. Herein, we take another pathway to attack black-box models in the fields of OD.

To launch a black-box attack using Daedalus, an ensemble of popular OD models is used as the substitute to generate adversarial examples. This heavily relies on the observed fact that current OD tasks tend to reuse few popular OD models with pre-trained backbones. As a consequence, the generated adversarial examples can be effective against all the popular models. Such a universal adversarial example has a high chance to break NMS of a black-box model. To find such an example, we can optimise a loss function as an expectation over OD models:

\begin{equation}\label{f4}
f = \displaystyle \mathop{\E}_{m\sim M}{f^m},
\end{equation}
herein, $M$ is a set of OD models with popular backbones, which includes the popular OD models and backbones. $f^m$ is the loss value of the $m$-th model in $M$, calculated based on one of Equations~\ref{f1},\ref{f2}, and \ref{f3}.

\subsection{The Daedalus attack against NMS}
In this section, the above steps are assembled to form the Daedalus attack. Actually, the attack can be applied to all FCN-based detectors. FCN-based detectors output the detected box dimension, the box position and the classification probabilities in a fully differentiable, end-to-end manner. Therefore, the attack can compute end-to-end adversarial gradients and optimise adversarial examples for FCN-based detectors.

\begin{algorithm}[h!]
\footnotesize
\caption{The $L_2$ Daedalus attack}
\label{A1}
\textbf{Input}: $x$, $\Lambda$, $\gamma$, $binary\_steps$, $\eta$, $max\_iteration$, $c_{max}$, $c_{min}$ \\
\textbf{Initialisation}: \\
$c \leftarrow 10$ \\
$loss_{init} \leftarrow f(x)$ \\
$\delta \leftarrow 0$ \\
$x^* \leftarrow x+\delta$ \\
\For{$n$ starts from 0 to $binary\_steps$}
{	
    \For{$i$ starts from 0 to $max\_iteration$}
    {
    		select boxes in $\Lambda$ to calculate loss $f(x^*)$\\
    		$\delta \leftarrow \delta - \eta  \bigtriangledown_{\delta}{[\|\delta\| + c \cdot f(x^*)]}$\\
    		$x^* \leftarrow x^*+\delta$
    }
    $x' \leftarrow$ the best $x^*$ found. \\
    \If{$f(x^*) <= loss_{init}\cdot (1-\gamma)$}
    {	
    		$c_{max} = min(c, c_{max})$\\
    		$c \leftarrow 0.5 \cdot (c_{max}+c_{min})$
    }
    \Else
    {
    		$c_{min} = max(c, c_{min})$\\
    		$c \leftarrow 0.5 \cdot (c_{max}+c_{min})$
    }
}
\textbf{Output}: Adversarial example $x'$.
\end{algorithm}

\begin{algorithm}[h!]
\footnotesize
\caption{The $L_0$ Daedalus attack}
\label{A2}
\textbf{Input}: $x$, $\Lambda$, $\gamma$, $binary\_steps$, $\eta$, $max\_iteration$, $c_{max}$, $c_{min}$ \\
\textbf{Initialisation}: \\
$c \leftarrow 10$ \\
$loss_{init} \leftarrow f(x)$ \\
$\delta \leftarrow 0$ \\
$x^* \leftarrow x+\delta$ \\
\For{$n$ starts from 0 to $binary\_steps$}
{
    \For{$i$ starts from 0 to $max\_iteration$}
    {
    		select boxes in $\Lambda$ to calculate loss $f(x^*)$\\
    		$\lambda = \argmax_{\lambda \in \delta} {\bigtriangledown_{\delta}{[\|\delta\| + c \cdot f(x^*)]}}$ \\
    		$\delta \leftarrow \delta - \eta \bigtriangledown_{\lambda}{[\|\delta\| + c \cdot f(x^*)]}$ \\
    		$x^* \leftarrow x^* + \delta$
    }
    $x' \leftarrow$ the best $x^*$ found. \\
    \If{$f(x^*) <= loss_{init}\cdot (1-\gamma)$}
    {
    		$c_{max} = min(c, c_{max})$ \\
    		$c \leftarrow 0.5 \cdot (c_{max}+c_{min})$
    }
    \Else
    {
    		$c_{min} = max(c, c_{min})$ \\
    		$c \leftarrow 0.5 \cdot (c_{max}+c_{min})$
    }
}
\textbf{Output}: Adversarial example $x'$.
\end{algorithm}

Given an OD task with an image $x$ as an input, the number of bounding boxes from NMS is maximised. As an example, YOLO-v3 outputs three feature maps that are in different scales. The three feature maps are concatenated into one final feature map. The details of the YOLO-v3 feature map are discussed in Section~\ref{YOLOv3}. We add a transformation layer after the feature map to obtain the final detection results. The transformation layer applies sigmoid transformations on the $t_x$ and $t_y$ in the feature map to get box centre coordinates $b_x$ and $b_y$. Exponential transformations are applied on $t_w$ and $t_h$ to obtain the box width $b_w$ and the box height $b_h$, respectively. We then calculate the values of the loss functions defined in Section~\ref{losses} and Section~\ref{ensemble_loss}. Next, we minimise the loss value together with the distortion to generate Daedalus adversarial examples. During the optimisation, we introduce a hyper-parameter $\gamma$ to control the strength of the attack.

The attack is bounded by two distortion metrics, namely $L_2$-norm and $L_0$-norm. Herein, an $L_p$-norm is defined as $L_p = (\sum_i^n{|\delta_i|^p})^{\frac{1}{p}}$, where $\delta_i$ is the perturbation on the $i$-th pixel. Therefore, an $L_2$-norm attack limits the perturbations added on all the pixels, while an $L_0$-norm attack limits the total number of pixels to be altered. We develop these two versions of attack to study the effectiveness of all-pixel perturbation and selective-pixel perturbation while generating Daedalus adversarial examples.

For the $L_2$ Daedalus attack, we can directly minimise Equation~\ref{obj_func}. However, in the $L_0$ Daedalus attack, since $L_0$-norm is a non-differentiable metric, we alternatively perturb the pixels that have the highest adversarial gradients in each iteration, and then other gradients are clipped to be 0. The detailed algorithm of the $L_2$ attack is presented in Algorithm~\ref{A1}, and the $L_0$ attack is presented in Algorithm~\ref{A2}. Notice that Daedalus only requires the box parameters to launch the attack. This means that the attack can be applied to the detection models that output feature maps which contain the geometric information of bounding boxes, even if they use different backbone networks.

\section{Evaluation}\label{Evaluation}
Extensive experiments are carried out to evaluate the performance of the Daedalus attack. The experiments are run on a machine with four RTX 2080ti GPUs and 128G of memory. The code of Daedalus is available on Github\footnote{\textit{https://github.com/NeuralSec/Daedalus-attack.}}. 

\subsection{Experiment settings}
In the experiments, YOLO-v3 is picked as the model to be attacked. A recommended objectness threshold of 0.5 is used in YOLO-v3. We first select the best loss function, and then we quantify the performance of the Daedalus attack based on YOLO-v3 with NMS. YOLO-v3 is employed as the substitute to generate $416\times 416$ adversarial perturbations in the evaluations. During the binary search to find the best constant $c$, the maximum number of search steps is set to $5$.

$L_0$ and $L_2$ Daedalus examples of images in the COCO 2017val dataset~\cite{lin2014microsoft} are crafted for the evaluation. For each attack, first, to investigate the effectiveness of different confidence values ($\gamma)$, we sweep $\gamma$ from $0.1$ to $0.9$ and perturb $10$ COCO examples under each $\gamma$. Second, $100$ COCO examples are randomly perturbed under both a low $(0.3)$ and a high $(0.7)$ value for $\gamma$. First, to unify the evaluation results, all of the $80$ object categories from the COCO dataset are included in the target set $\Lambda$ to be attacked. Second, the categories of `person' and `car' are selected to be attacked. The graphical results are included in the supplementary file of the paper.

To prove that the attack can be applied on other detectors, the results of attacking RetinaNet-ResNet-50 and SSD are also presented. RetinaNet-ResNet-50 uses ResNet-50~\cite{he2016deep} as the backbone. SSD employs VGG-16 as its backbone~\cite{liu2016ssd}. The backbones differ from that of YOLO-v3. The output feature maps are different as well. An ensemble of the three OD models are used as the substitute to craft universal adversarial examples. Furthermore, to demonstrate the effectiveness of the attack in the real world, Daedalus perturbations are made into physical posters to launch attacks against real-world OD applications. As a showcase, we show the success of the physical Daedalus attack in an indoor OD scene.

\subsection{Loss function selection}
The proposed adversarial loss functions (Equations~\ref{f1}, \ref{f2}, and \ref{f3}) are compared in this section. We select ten random images from the COCO dataset. For each function, adversarial examples of these ten images are generated. We record the runtime per example and the volatile GPU utilisation of each loss function. The performance of each function is measured by looking at the success rate of finding adversarial examples at different confidence scores. Since the $L_0$ attack essentially has the same optimisation process as the $L_2$ does, we only run the $L_2$ attack for the comparison purpose. The results are reported in Table~\ref{function_comparison}.

The recorded runtime is averaged over the runtime of generating all examples under the same loss function. The GPU utilisation is also averaged over all examples. We use only one GPU to generate an adversarial example in this section. It can be observed that $f_3$ (\textit{i.e.}, Equation~\ref{f3}) has the best success rate of finding adversarial examples under different $\gamma$ values. Moreover, $f_3$ has the lowest runtime as well as the lowest GPU utilisation compared to $f_1$ and $f_2$.

Based on the comparisons, $f_3$ is selected as the loss function for Daedalus. The Daedalus attack is evaluated in the following sections with $f_3$.

\begin{table*}[t!]
\caption{Loss function comparison}
\label{function_comparison}
\centering
\resizebox{0.65\linewidth}{!}{%
\begin{tabular}{|c|c|c|c|c|c|c|c|c|c|c|}
\hline
\multirow{2}{*}{\textit{\textbf{Loss Function}}} & \multirow{2}{*}{\textit{\textbf{Metric}}} & \multicolumn{9}{c|}{\textit{\textbf{Value of $\gamma$}}} \\ \cline{3-11} 
 &  & \multicolumn{1}{l|}{0.1} & \multicolumn{1}{l|}{0.2} & \multicolumn{1}{l|}{0.3} & \multicolumn{1}{l|}{0.4} & \multicolumn{1}{l|}{0.5} & \multicolumn{1}{l|}{0.6} & \multicolumn{1}{l|}{0.7} & \multicolumn{1}{l|}{0.8} & \multicolumn{1}{l|}{0.9} \\ \hline
\multirow{3}{*}{\textit{$f_1$}} & \textit{Runtime} & \multicolumn{9}{c|}{2375.62 s} \\ \cline{2-11} 
 & \textit{GPU-Util} & \multicolumn{9}{c|}{79\%} \\ \cline{2-11} 
 & \textit{Success rate} & \textbf{100\%} & \textbf{100\%} & \textbf{100\%} & 90\% & 40\% & 40\% & 20\% & 0\% & 0\% \\ \hline
\multirow{3}{*}{\textit{$f_2$}} & \textit{Runtime} & \multicolumn{9}{c|}{1548.57 s} \\ \cline{2-11} 
 & \textit{GPU-Util} & \multicolumn{9}{c|}{77\%} \\ \cline{2-11} 
 & \textit{Success rate} & \textbf{100\%} & \textbf{100\%} & \textbf{100\%} & \textbf{100\%} & \textbf{100\%} & 90\% & 40\% & 20\% & 0\% \\ \hline
\multirow{3}{*}{\textit{$f_3$}} & \textit{Runtime} & \multicolumn{9}{c|}{1337.06 s} \\ \cline{2-11} 
 & \textit{GPU-Util} & \multicolumn{9}{c|}{73\%} \\ \cline{2-11} 
 & \textit{Success rate} & \textbf{100\%} & \textbf{100\%} & \textbf{100\%} & \textbf{100\%} & \textbf{100\%} & \textbf{100\%} & 70\% & 40\% & 40\% \\ \hline
\end{tabular}
}
\end{table*}

\subsection{Quantitative performance}
In this section, the attack is evaluated from two perspectives. First, we evaluate how the Daedalus attack performs under different NMS thresholds. Second, we investigate how the attack confidence level ($\gamma$) affects detection results. The performance of the attack is assessed on a YOLO-v3 detector. The attack performance is quantified based on three metrics: the False Positive ($FP$) rate, the mean Average Precision ($mAP$), and the distortion of the example.

First, to investigate the effect of NMS threshold on the attack, we obtain the detection results of each adversarial/benign example under a series of NMS thresholds, which range between $0.5$ and $0.95$. Second, to assess the effect of confidence value ($\gamma$) on attack performance, adversarial examples are crafted for confidences from $0.1$ to $0.9$.

\noindent
\textit{\textbf{False positive rate.}} The $FP$ rate is defined as the ratio of false positive detection boxes with respect to the total number of detection boxes.

\begin{equation}
FP\ rate = \frac{N_{\phi}}{N} ,
\end{equation}
wherein $N_{\phi}$ is the number of false positive detection boxes, and $N$ is the total number of bounding boxes found by the model. $FP$ rate measures the proportion of redundant boxes that are not filtered by NMS. $FP$ is maximally one and that is when there is no correct detection result.
A higher value of $FP$ indicates a higher success rate for the attack.
\begin{figure}[t]
\center
\includegraphics[width=0.85\linewidth]{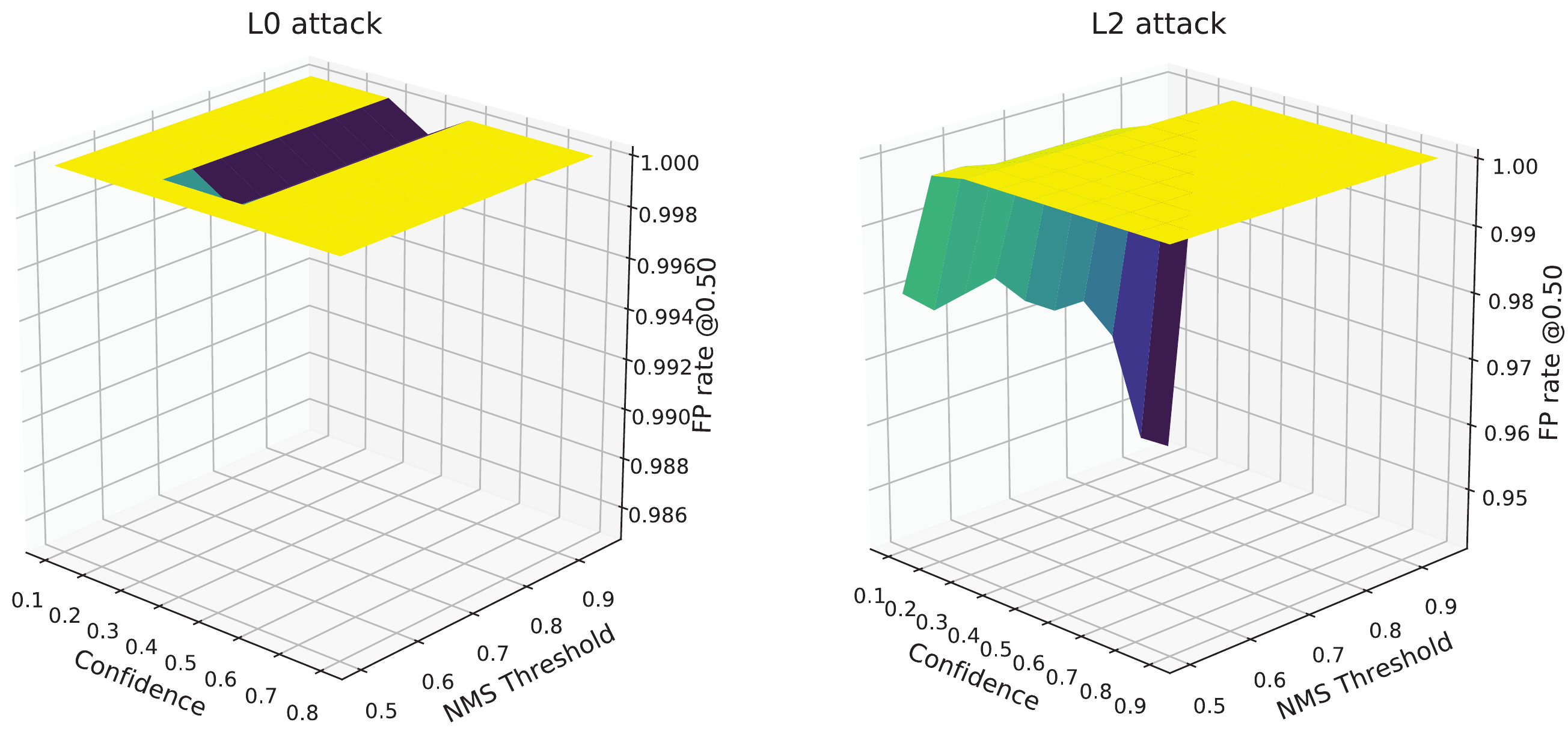}
\caption{The false positive rate at IoU threshold of 0.5 with respect to NMS threshold and confidence of attack. The NMS threshold ranges from $0.5$ to $0.95$. The confidence ranges from $0.1$ to $0.9$ in $L_2$ attack, $0.1$ to $0.8$ in $L_0$ attack.}
\label{FP_rate_50}
\end{figure}

\begin{figure}[t]
\center
\includegraphics[width=0.85\linewidth]{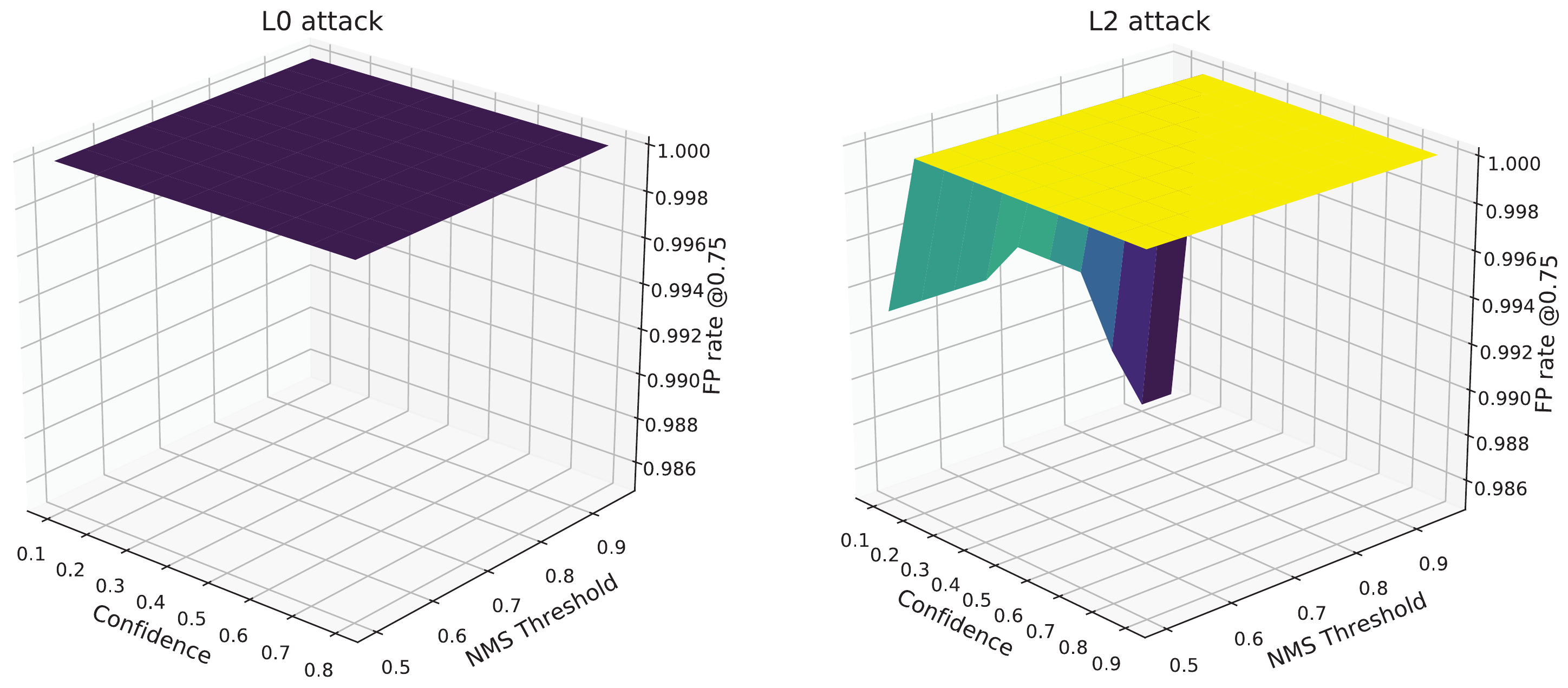}
\caption{The false positive rate at IoU threshold of 0.75 with respect to NMS threshold and confidence of attack. The NMS threshold ranges from $0.5$ to $0.95$. The confidence ranges from $0.1$ to $0.9$ in $L_2$ attack, $0.1$ to $0.8$ in $L_0$ attack.}
\label{FP_rate_75}
\end{figure}

\begin{figure}[t]
\center
\includegraphics[width=0.85\linewidth]{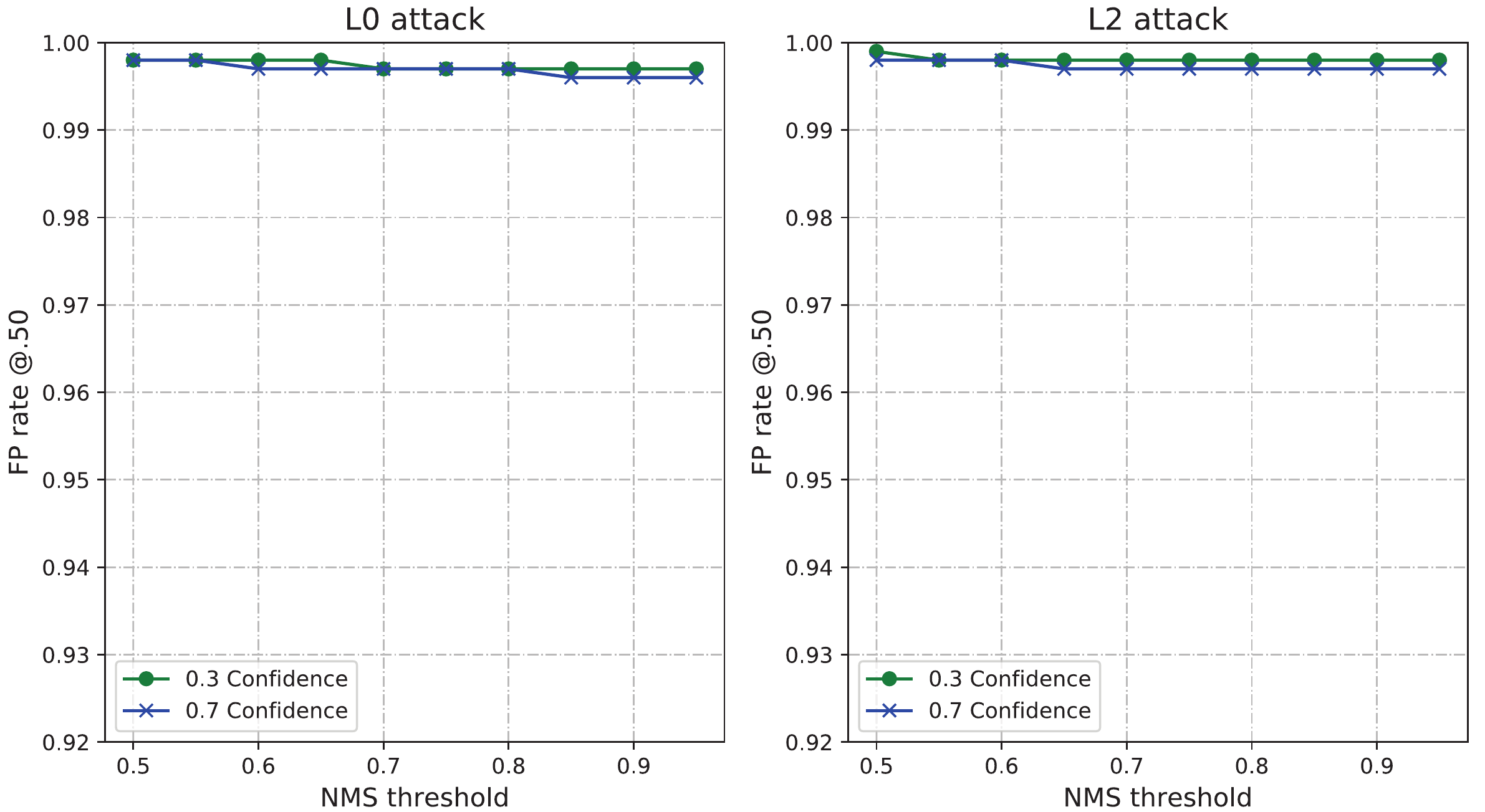}
\caption{$L_0$ attack false positive rate at IoU threshold of 0.5 with respect to NMS threshold. Each $FP$ rate is averaged from the detection results of $100$ Daedalus examples.}
\label{FP_rate_l00_50}
\end{figure}

\begin{figure}[t]
\center
\includegraphics[width=0.85\linewidth]{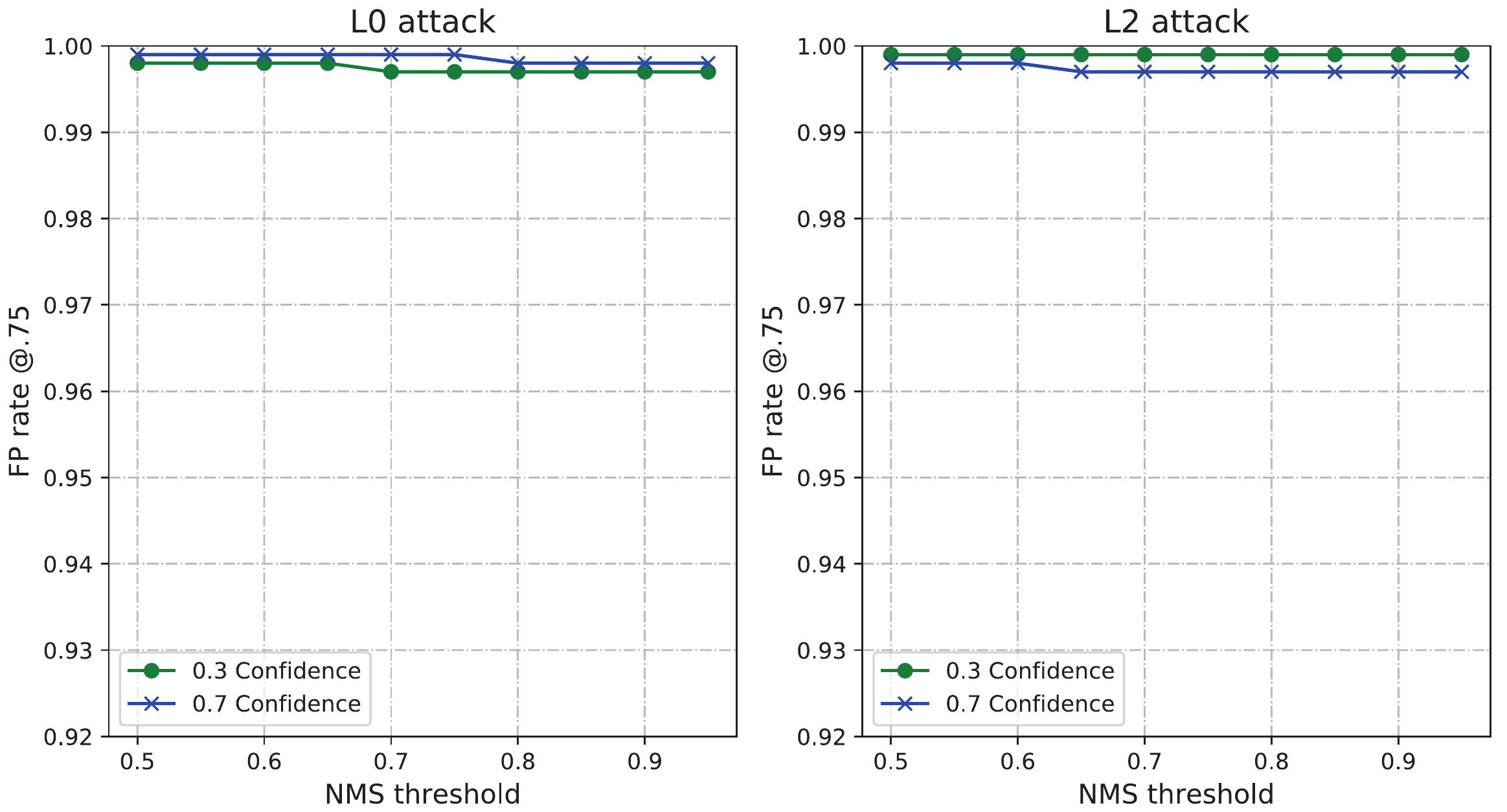}
\caption{$L_0$ attack false positive rate at IoU threshold of 0.75 with respect to NMS threshold. Each $FP$ rate is averaged from the detection results of $100$ Daedalus examples.}
\label{FP_rate_l00_75}
\end{figure}

The IoU threshold in NMS is varied from $0.5$ to $0.95$ with steps of $0.05$ to obtain detection outputs under different thresholds. The detection results of Daedalus examples made under confidences from $0.1$ to $0.9$ ($0.1$ to $0.8$ for $L_0$ since it cannot find examples with $0.9$ confidence) is then obtained. This is because the $L_0$ cannot further discard perturbations from the $L_2$ examples under a high $\gamma$. We have reported the $FP$ rates in the results based on the IoU thresholds of $0.5$ and $0.75$ in Fig.~\ref{FP_rate_50} and Fig.~\ref{FP_rate_75}, respectively. Each $FP$ rate is averaged over $10$ Daedalus examples. Next, we measure the average $FP$ rate of $100$ examples crafted under a confidence value of $0.3$/$0.7$. The trends of the average $FP$ rate at IoU thresholds of $0.5$ and $0.75$ with respect to the NMS threshold are plotted in Fig.~\ref{FP_rate_l00_50} and Fig.~\ref{FP_rate_l00_75}, respectively.

It can be observed that both $L_0$ and $L_2$ attacks achieve $FP$ rates of at least $90\%$ under different NMS thresholds, even with the lowest attack confidence of $0.1$. Increasing the NMS threshold slightly reduces the $FP$ rate of low-confidence attacks. However, it barely has an impact on the performance of high-confidence attacks by changing the NMS threshold. $FP$ rate increases with the increase of $\gamma$. It can get as high as $99.9\%$ for a high-confidence Daedalus attack, even with a stricter NMS threshold.

\noindent
\textit{\textbf{Mean average precision.}} The $mAP$ is commonly used to evaluate the performance of OD model. It averages the Average Precisions ($AP$s) over all detected object categories. This metric is defined as follows:

\begin{equation}
mAP = \frac{1}{M}\sum_{i=1}^M{AP_i},
\end{equation}
in which $M$ is the number of object categories. We adopted the all-point interpolation $AP$ from Pascal VOC challenge~\cite{everingham2010pascal} to calculate $mAP$.

\begin{figure}[t]
\center
\includegraphics[width=0.85\linewidth]{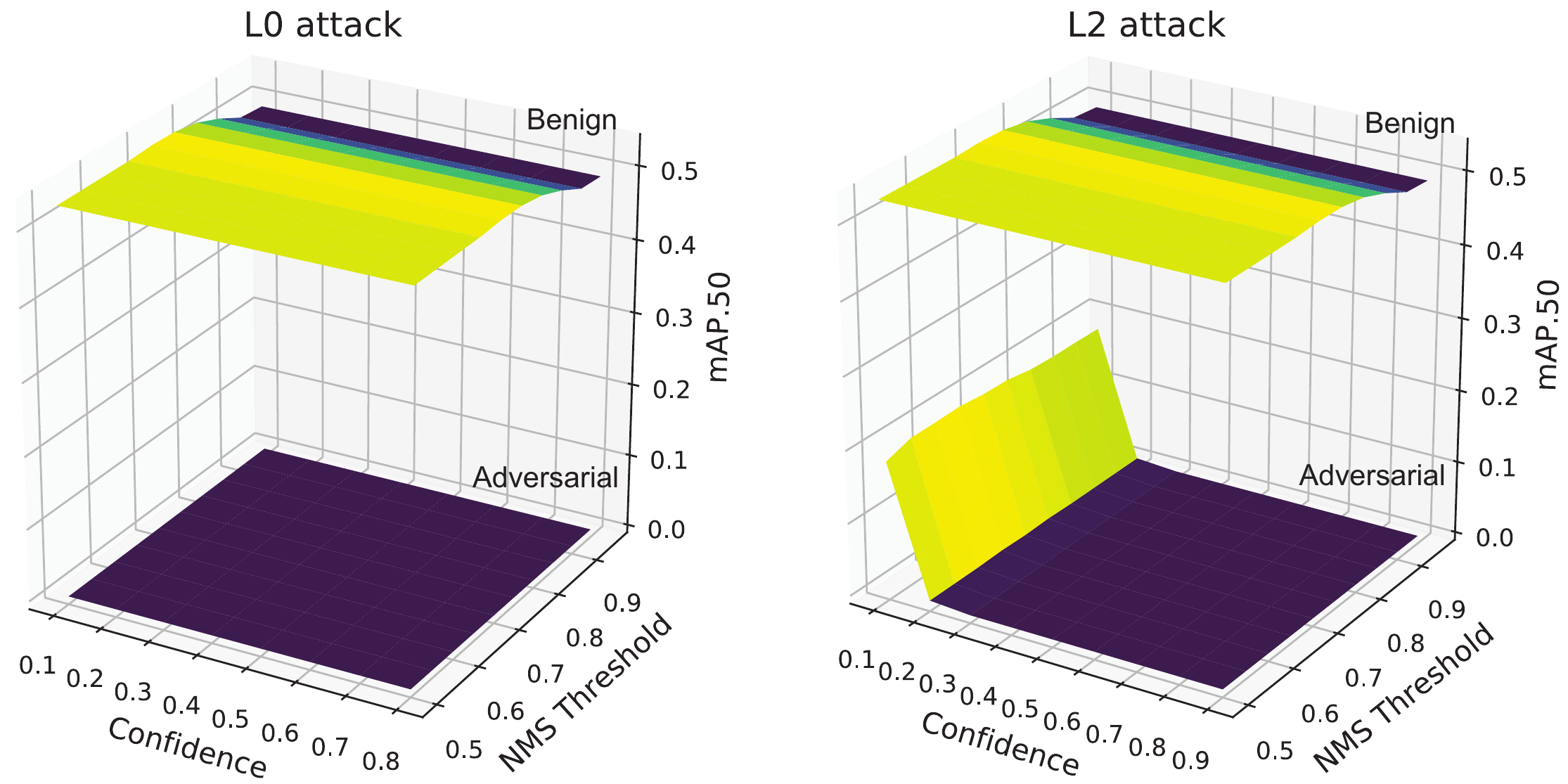}
\caption{$mAP^{IoU=.50}$ of Daedalus example detection results with respect to NMS threshold and attack confidence. The threshold ranges from $0.5$ to $0.95$.}
\label{mAP50}
\end{figure}

\begin{figure}[t]
\center
\includegraphics[width=0.85\linewidth]{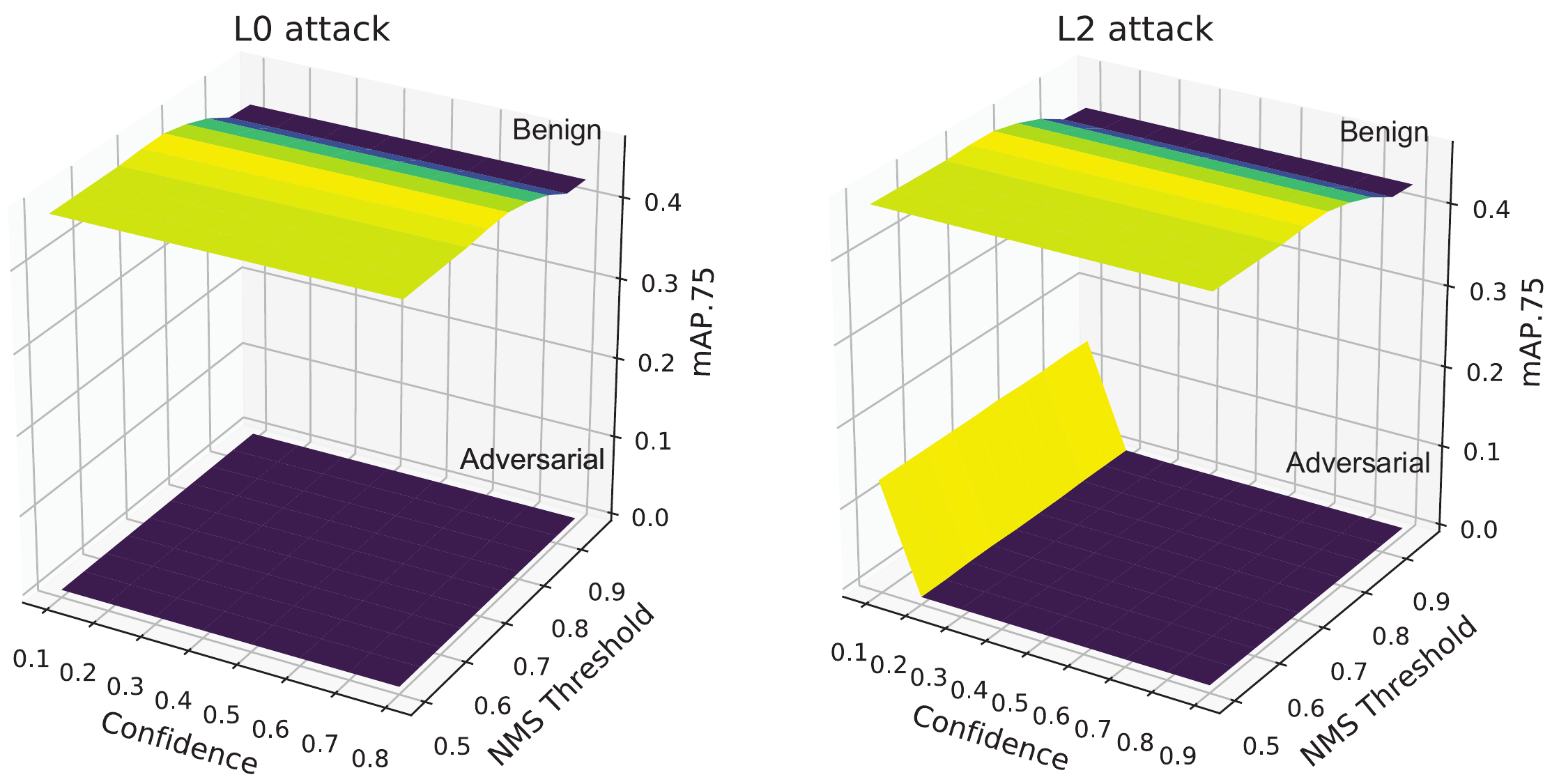}
\caption{$mAP^{IoU=.75}$ of Daedalus example detection results with respect to NMS threshold and attack confidence. The threshold ranges from $0.5$ to $0.95$.}
\label{mAP75}
\end{figure}

\begin{figure}[t]
\center
\includegraphics[width=0.85\linewidth]{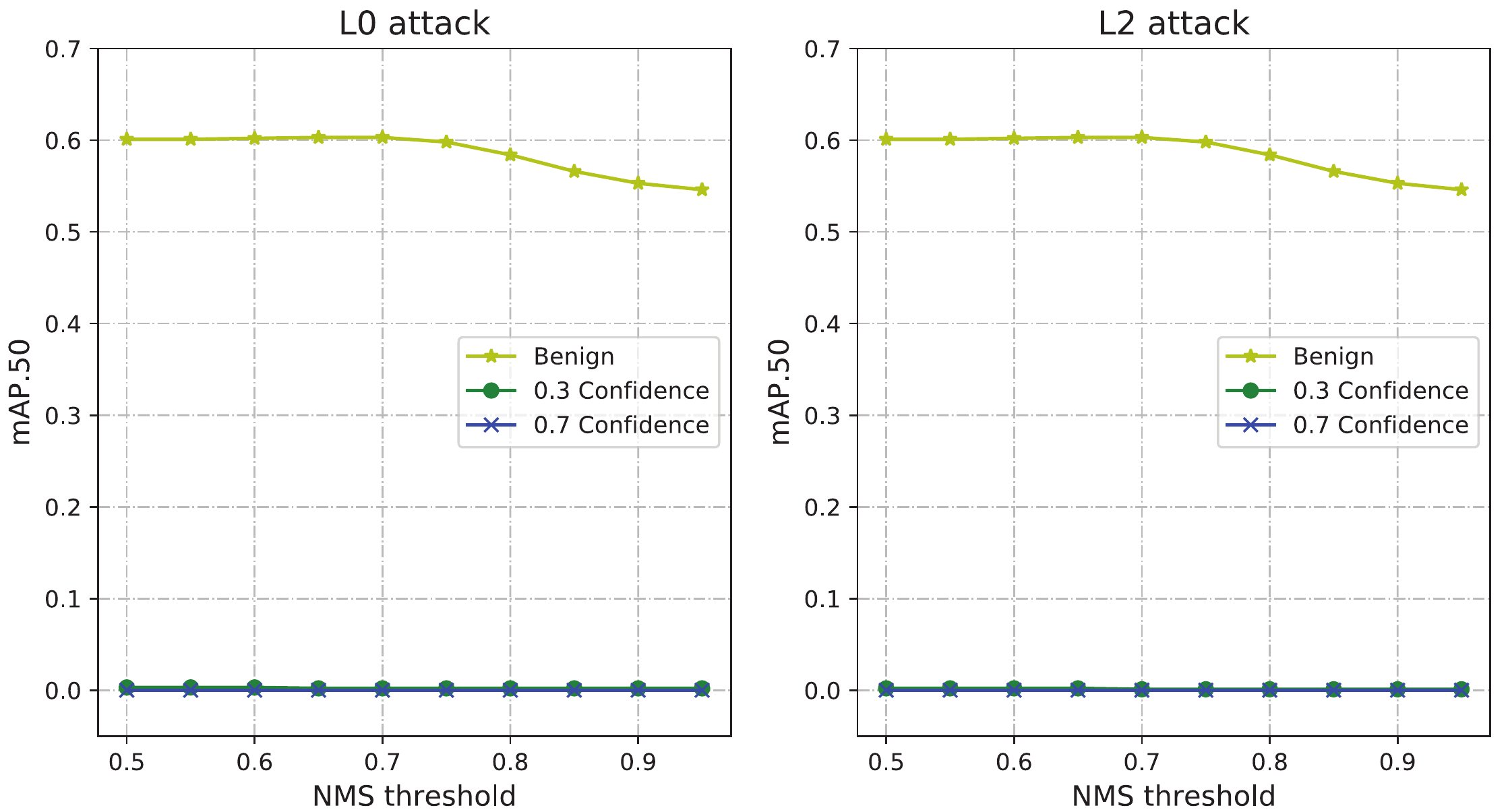}
\caption{$mAP^{IoU=.50}$ of the $L_0$ and $L_2$ Daedalus attacks with respect to NMS threshold. The threshold ranges from $0.5$ to $0.95$. Each mAP is averaged from the detection results of $100$ Daedalus examples.}
\label{mAP50_100}
\end{figure}

\begin{figure}[t]
\center
\includegraphics[width=0.85\linewidth]{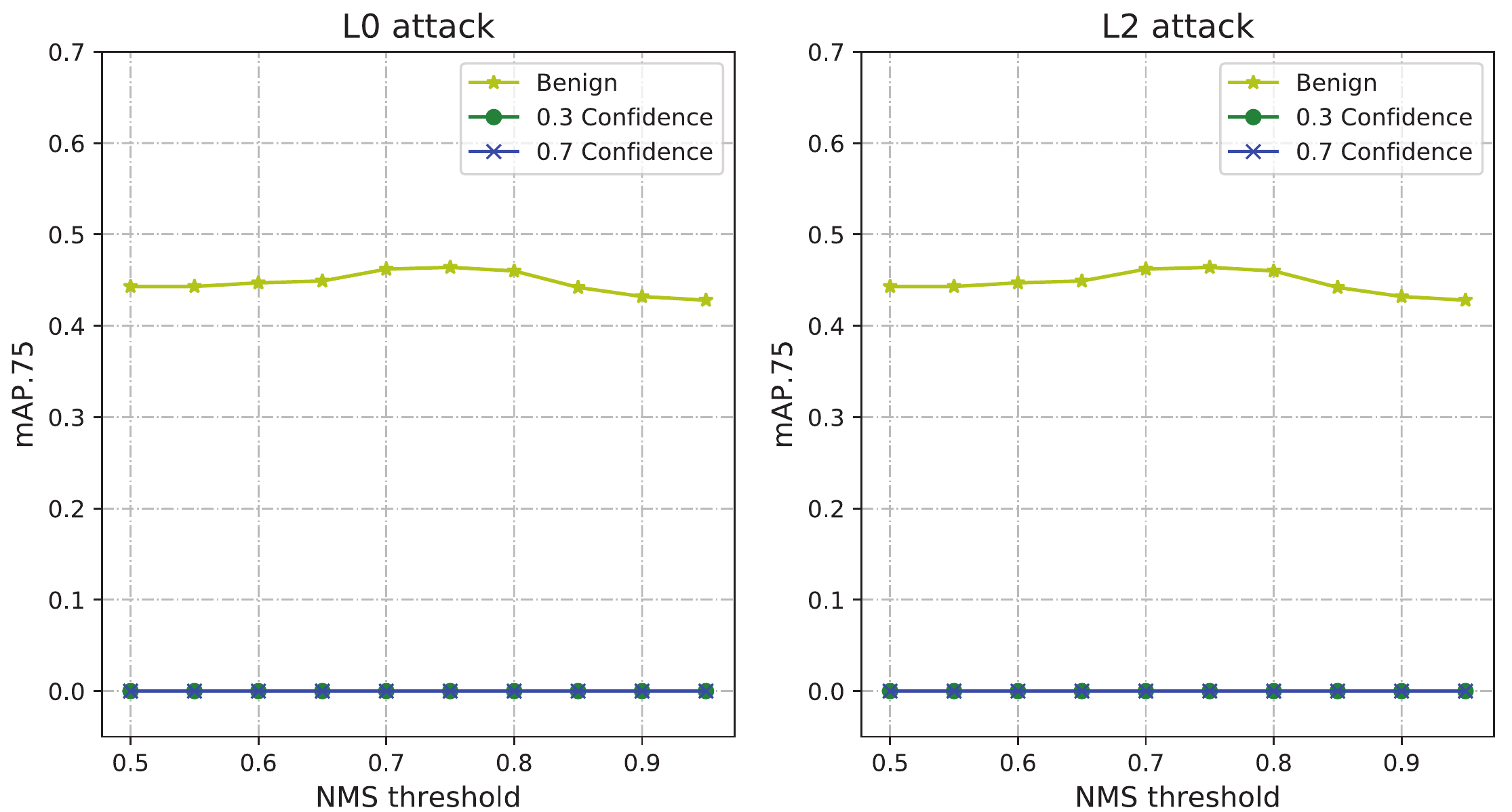}
\caption{$mAP^{IoU=.75}$ of the $L_0$ and $L_2$ Daedalus attacks with respect to NMS threshold. The threshold ranges from $0.5$ to $0.95$. Each mAP is averaged from the detection results of $100$ Daedalus examples.}
\label{mAP75_100}
\end{figure}

The recommended COCO evaluation metrics $mAP^{IoU=.50}$ and $mAP^{IoU=.75}$, which are the $mAP$ values calculated at IoU thresholds of $0.5$ and $0.75$, respectively, are adopted to measure the detection precisions on benign examples and Daedalus examples. Fig.~\ref{mAP50} and Fig.~\ref{mAP75} show the $mAP$ values of the detection results with changing NMS threshold and $\gamma$. We also plotted the average $mAP^{IoU=.50}$ and $mAP^{IoU=.75}$ for $100$ examples made under confidence values of $0.3$ and $0.7$ in Fig.~\ref{mAP50_100} and Fig.~\ref{mAP75_100}, respectively.

Based on the results, both $L_0$ and $L_2$ attacks decrease $mAP$ of the detector from $45\% \sim 59\%$ to a value between $0\%$ to $17.8\%$. $mAP$ drops to $0\%$ when the confidence of the attack surpasses $0.2$.

\noindent
\textit{\textbf{Distortion of the examples.}} The distortions of the Daedalus adversarial examples crafted based on $L_2$-norm and $L_0$-norm are evaluated in this part. The maximum, the minimum and the average distortion of $10$ adversarial examples are recorded for each attack confidence value. The results are summarised in Table~\ref{dist}. It can be seen that the distortion of the adversarial examples has a positive correlation with $\gamma$. The distortion increases slowly before the confidence reaches $0.6$. Once the confidence goes beyond $0.6$, there is a rapid increase in the distortion amount. Additionally, compared to the $L_2$ attack, the $L_0$ one cannot find any adversarial examples at the confidence of $0.9$.

\begin{table*}[h!]
\caption{Distortions under different confidences}
\label{dist}
\centering
\resizebox{0.75\linewidth}{!}{%
\begin{tabular}{|c|c|c c c c c c c c c|}
\hline
\multirow{2}{*}{\textit{\textbf{Attack}}} & \multirow{2}{*}{\textit{\textbf{Distortion}}} & \multicolumn{9}{c|}{\textit{\textbf{Value of \textbf{$\gamma$}}}} \\ 
 &  & 0.1 & 0.2 & 0.3 & 0.4 & 0.5 & 0.6 & 0.7 & 0.8 & 0.9 \\ \hline
\multirow{3}{*}{\textit{$L_2$ Daedalus}} & \textit{Maximum} & 157.49 & 259.25 & 113.24 & 324.53 & 598.45 & 4138.47 & 5915.90 & 25859.67 & 19553.05 \\ \cline{2-11} 
 & \textit{Average} & 130.59 & 218.86 & 96.51 & 166.27 & 344.75 & 1670.35 & 1835.04 & 9953.59 & 14390.76 \\ \cline{2-11} 
 & \textit{Minimum} & 94.30 & 175.07 & 75.24 & 95.30 & 185.25 & 351.55 & 366.49 & 210.24 & 11894.38 \\ \hline
\multirow{3}{*}{\textit{$L_0$ Daedalus}} & \textit{Maximum} & 162.74 & 513.54 & 555.11 & 1200.48 & 1001.86 & 42144.96 & 3891.80 & 3364.29 & N/A \\ \cline{2-11}
 & \textit{Average} & 87.81 & 346.71 & 440.21 & 651.77 & 736.40 & 12765.82 & 1992.30 & 1501.58 & N/A \\ \cline{2-11} 
 & \textit{Minimum} & 53.26 & 165.50 & 339.44 & 402.48 & 436.15 & 3713.67 & 729.38 & 848.36 & N/A \\ \hline
\end{tabular}
}
\end{table*}

Based on the evaluation, both $L_0$ and $L_2$ attacks require moderate distortions to make low-confidence examples. Compared to the $L_0$ attack, the $L_2$ attack introduces less distortion when the confidence is between $0.2$ and $0.7$, and it is slightly better in finding adversarial examples with the confidence of $0.9$. However,the $L_0$ attack can reduce the distortion when the confidence goes above $0.8$. It is known that FCN-based object detectors usually segment the input image by a mesh grid. Each grid cell has its own responsibility for detecting objects. Low-level perturbation based on $L_0$-norm may leave some of the grid cells unperturbed. Therefore, these cells can still output normal detection results, which can be filtered by NMS. As a result, the $L_0$ attack perturbs most of the pixels to generate a well-qualified adversarial example, which implies having  high distortions in the example. High-level perturbation is imposed on each pixel when the confidence goes above $0.8$. However, the $L_0$ attack can discard the perturbation on some pixels to reduce the overall distortion. 

\section{Analysis of the attack}\label{Discussion}
In this section, we first investigate whether the Daedalus attack can be extended to a variation of NMS algorithm, namely soft-NMS~\cite{bodla2017soft} or not. Second, we demonstrate the universality of the attack by applying it to attack RetinaNet-Resnet-50. Third, an \textit{ensemble of substitutes} is proposed to create universal Daedalus examples that can attack multiple object detectors. The proposed ensemble attack makes Daedalus examples transferable so that they can be used in black-box scenarios. Subsequently, posters made of Daedalus perturbations are used to launch attacks against real-world OD systems. Finally, we discuss the problem of defending against the attack. 

\subsection{Breaking Soft-NMS}
\begin{figure}[htb]
\centering
  \begin{subfigure}[b]{.46\linewidth}
    \centering
    \includegraphics[width=1\linewidth]{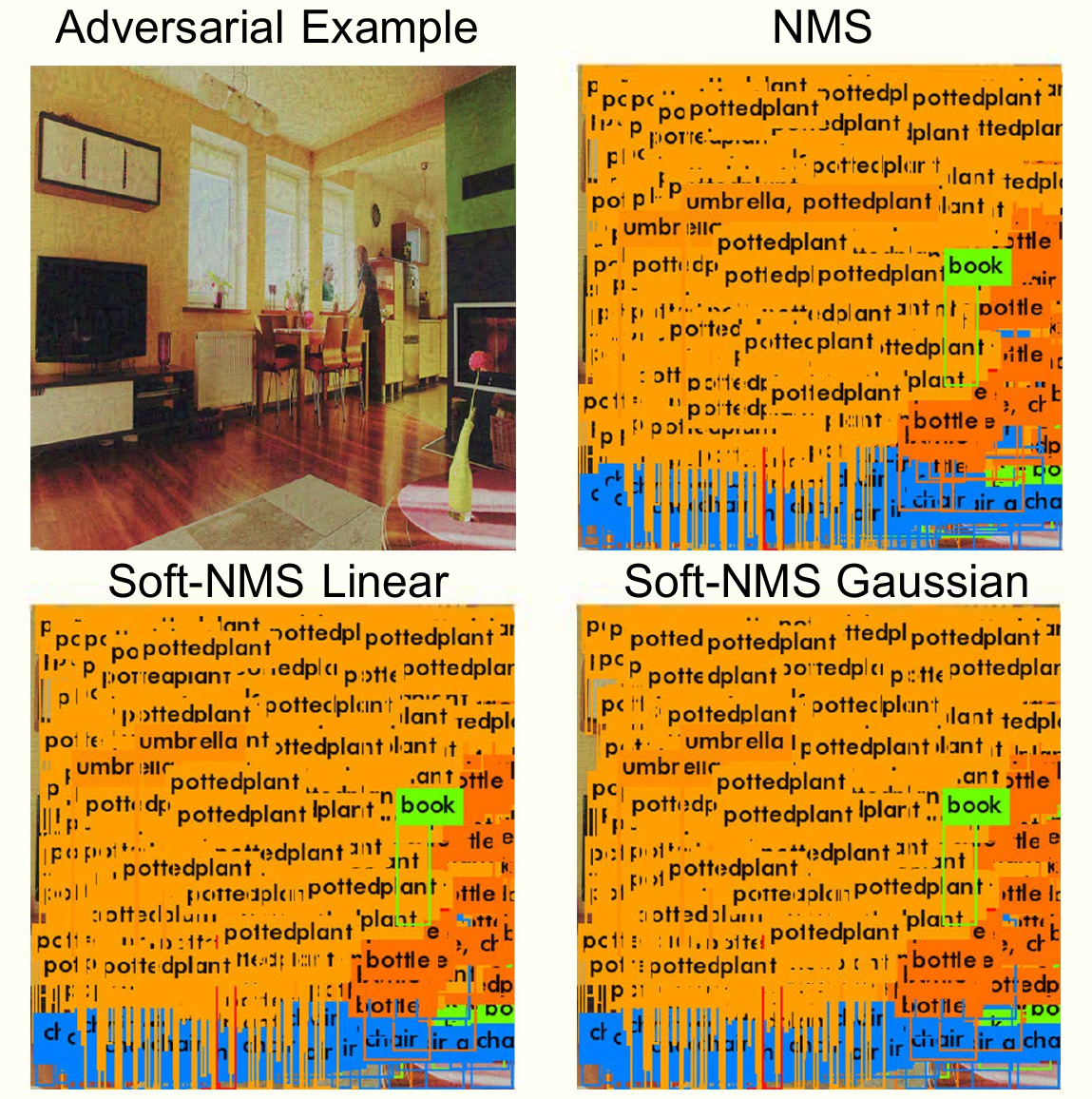}
	\caption{}
	\label{softnms_L0}
  \end{subfigure}%
  \rulesep
  \begin{subfigure}[b]{.45\linewidth}
    \centering
    \includegraphics[width=0.98\linewidth]{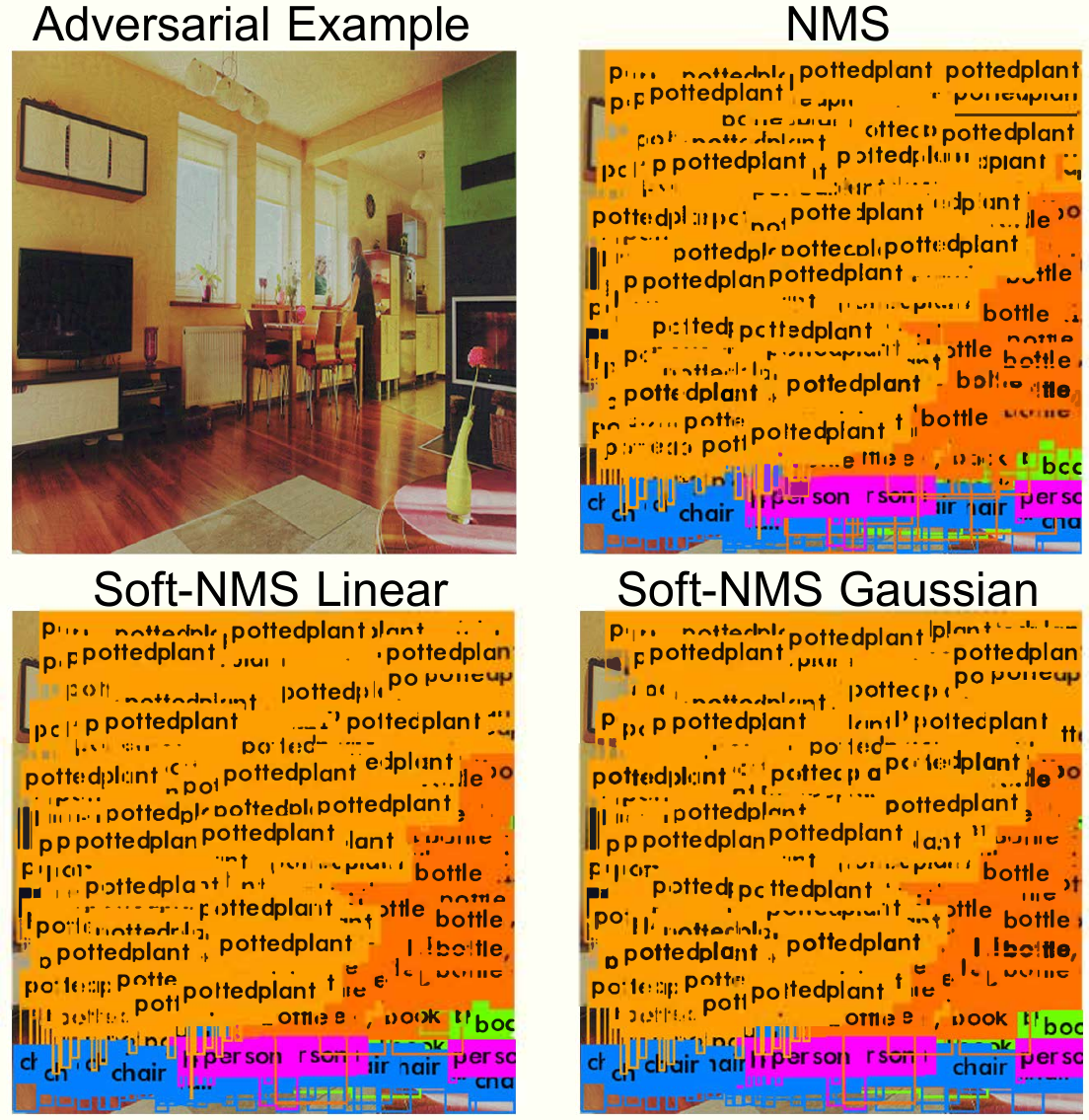}
    \caption{}
    \label{softnms_L2}
  \end{subfigure}%
\caption{(a) The $L_0$ Daedalus attack against soft-NMS. The top left image is an $L_0$ Daedalus example crafted under a confidence of $0.3$. The top right one shows the detection results from NMS, as a reference. The left bottom one shows the results of linear soft-NMS, and the right bottom one shows the results of Gaussian soft-NMS. (b) The $L_2$ Daedalus attack against soft-NMS. The top left image is an $L_2$ Daedalus example crafted under a confidence of $0.3$. The top right one shows the detection results from NMS, as a reference. The left bottom one shows the results of linear soft-NMS, and the right bottom one shows the results of Gaussian soft-NMS.}
\end{figure}

In addition to breaking NMS, the Daedalus attack is tested against soft-NMS, which iteratively decreases the confidence score of overlapping boxes instead of discarding them~\cite{bodla2017soft}. Soft-NMS has linear and Gaussian versions. We attack both versions to investigate whether soft-NMS can handle it or not. We randomly select an image to craft its $L_0$ and $L_2$ Daedalus examples to attack a soft-NMS-equipped YOLO-v3 detector. The detection results for $L_0$ adversarial examples are displayed in Fig.~\ref{softnms_L0}. For $L_2$ adversarial examples, the detection results are similarly illustrated in Fig.~\ref{softnms_L2}. From the results one can find that both  $L_0$ and $L_2$ can break linear soft-NMS as well as Gaussian soft-NMS.

\begin{figure}[t]
\center
\includegraphics[width=0.85\linewidth]{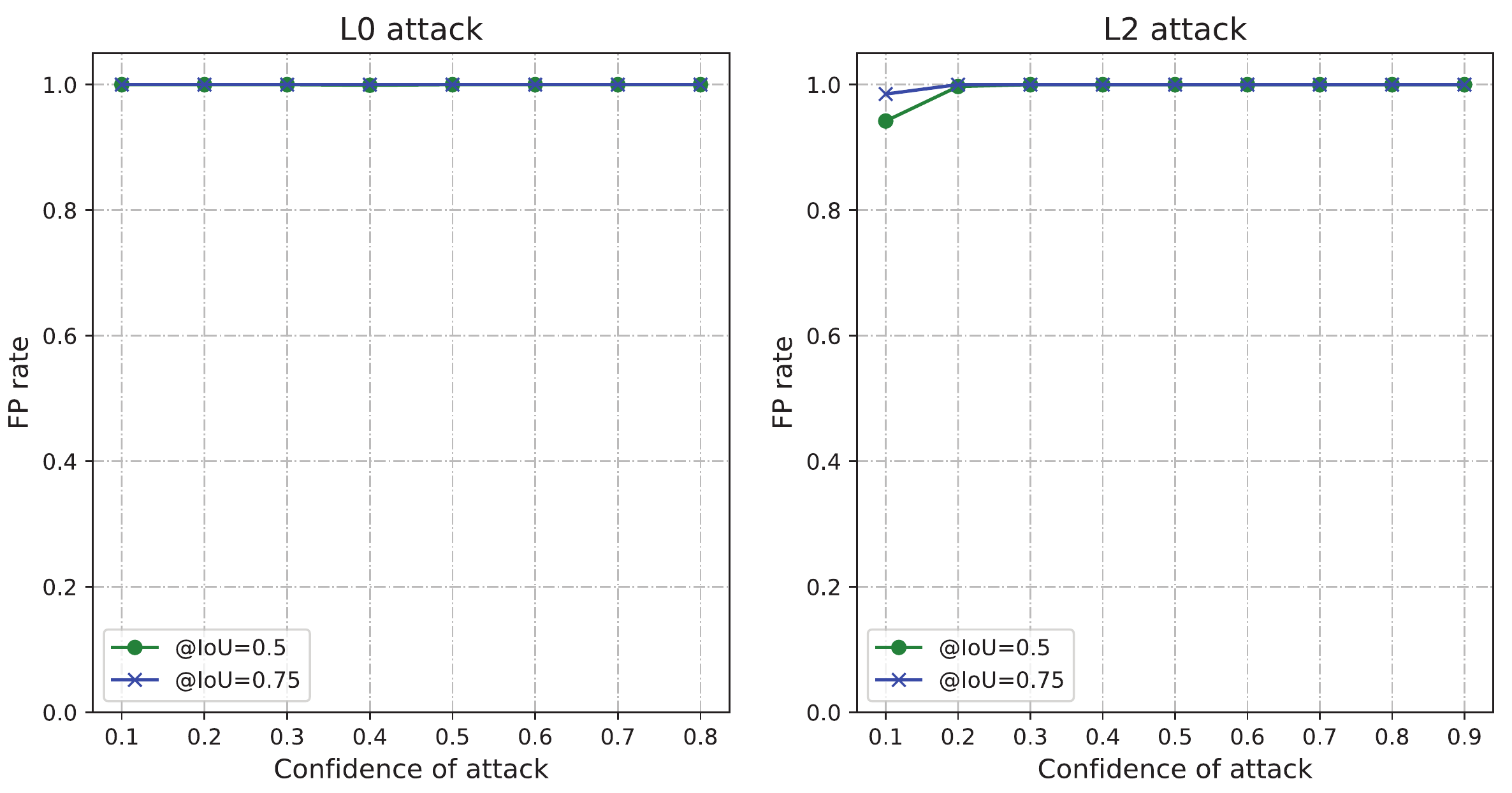}
\caption{The $FP$ rate of the Daedalus attack against YOLO-v3 with soft-NMS. We evaluated the FP rate at IoU thresholds of $0.5$ and $0.75$.}
\label{softnms_FP}
\end{figure}

The FP rate of the attack for soft-NMS is also measured. The results are shown in Fig.~\ref{softnms_FP}. The $L_0$ adversarial examples which are made with different confidence values result in a $FP$ rate of $100\%$. The $L_2$ attack also leads to $100\%$ $FP$ rate when the attack confidence is above $0.2$.

\subsection{Universal Daedalus examples}
Based on the census in Section~\ref{census}, three popular models, namely YOLO-v3, SSD, and RetinaNet, are employed to verify the idea of ensemble attack. We do not introduce more models as substitutes, simply due to the limited GPU memory. According to Table~\ref{census}, a universal Daedalus example for YOLO-v3, SSD, and RetinaNet can already break $54.9\%$ of the 1,696 repositories. We investigate whether a universal Daedalus example can be found to break all the models. In the experiments, YOLO-v3 employs Darknet-53 as its backbone network. RetinaNet and SSD have different backbone networks and use ResNet-50 and VGG-16, respectively. Since the three models are backed by different backbones, they generate divergent feature maps.

\begin{figure}[t]
\center
\includegraphics[width=1\linewidth]{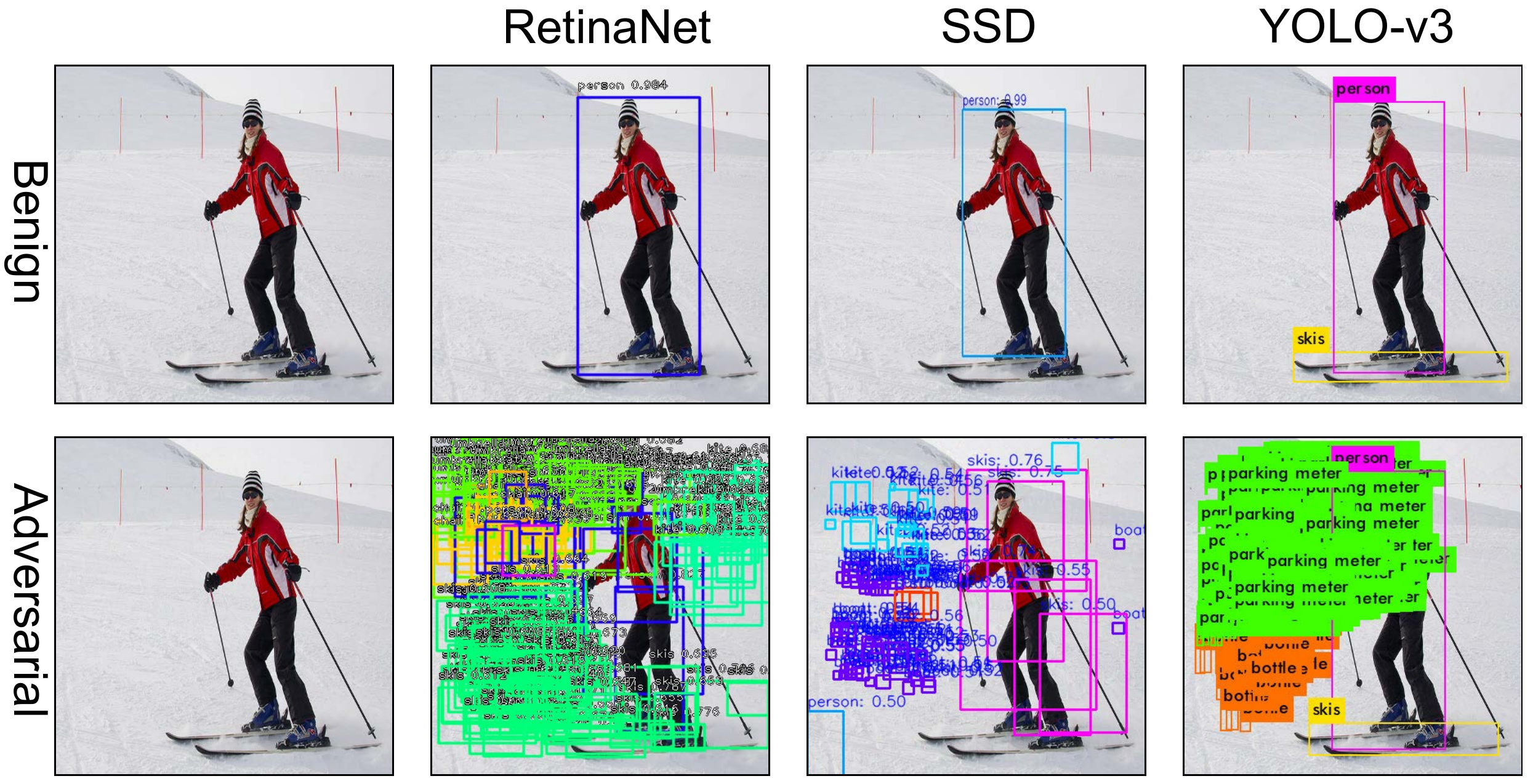}
\caption{The detection results of a Daedalus adversarial example made by ensemble-of-substitutes approach. We detect the adversarial example using YOLO-v3, RetinaNet, and SSD. The first row contains a benign example and its detection results. The second row contains an adversarial example and the detection results of the adversarial example.}
\label{ensemble_demo}
\end{figure}

The detection results of the adversarial example are displayed in Fig.~\ref{ensemble_demo}. It can be observed that the adversarial example can trigger NMS malfunctioning for all the models. The detection results for more adversarial examples crafted using the \textit{ensemble of substitutes} can be found in the supplementary file of this paper. Furthermore, when there is a large number of OD models in the ensemble, an attacker may adopt methods such as \cite{sener2018multi} to find proper coefficient for each term in the loss function to enhance the attack performance.

\subsection{Physical Daedalus examples}
To launch the Daedalus attack in the real world, the perturbation found by the attack can be instantiated into a poster. When the poster is captured by the camera of an object detection system, it will trigger the NMS malfunctioning. Given a scene to attack, we first film a video of the scene. The video contains a set $X$ of image frames. Next, we can optimise a poster of the perturbation $\delta$ as follows:

\begin{equation}\label{physical_loss}
\argmin_{\delta}\E_{x\sim X}\E_{\phi}{f_{3}(x, \delta, \Phi)} + SNPS(\delta, \beta)
\end{equation}

Herein, $\Phi$ is a set of random transformations (\textit{e.g.}, zooming, rotation, \textit{etc.}). The detailed transformations are summarised in the supplementary file. SNPS is a \textit{Sub-sampled Non-Printability Score} of the poster $\delta$. Non-Printability Score (NPS) is used for measuring the error between a printed pixel and its digital counterparts. Sum of NPSs of all pixels in an image can be used to measure the image printability~\cite{sharif2016accessorize}. However, it is expensive to compute the NPS of a large image. Therefore, given a sample rate $\beta$ and pixels of $\delta$, we sample a subset of the pixels to compute the SNPS. It is found that calculating the NPS from only $0.1\%\sim 1\%$ of the pixels is sufficient for making an effective Daedalus poster. Moreover, it is observed that the poster can be generalised across different scenes. 

Next, the robustness of three different sized (\textit{i.e.},$400\times 400$, $720\times 720$, and $1280\times 1280$) Daedalus posters are evaluated in an indoor environment. The effectiveness of the posters is recorded in Table~\ref{poster_val}.

\begin{table*}[t]
\caption{Effectiveness of the Daedalus posters}
\label{poster_val}
\centering
\begin{tabular}{c|c|c|c}
\hline
Metric & \multicolumn{3}{c}{Effectiveness}\\
\hline
& Small poster ($400\times 400$) & Medium poster ($720\times 720$) & Large poster ($1280\times 1280$) \\
\hline
Horizontal angle & $-16.5^{\circ} \sim 16.5^{\circ}$ & $-18^{\circ} \sim 18^{\circ}$ & $-19.5^{\circ} \sim 19.5^{\circ}$\\
Vertical angle   & $-11.5^{\circ} \sim 11.5^{\circ}$ & $-12^{\circ} \sim 12^{\circ}$ & $-12^{\circ} \sim 12^{\circ}$\\
Distance         & $0.12 \sim 1.2\ metres$ & $0.21 \sim 1.5\ metres$ & $0.2 \sim 1.8\ metres$\\                                            
\hline
\end{tabular}
\end{table*}

In the scenario of Smart Inventory Management (SIM), an OD model will be used to identify the items that are placed into or withdrawn from an inventory. We simulate an attack against the case of depositing a MacBook and a monitor into the inventory. A demonstration of the attack can be found in the supplementary file.

\subsection{Adaptive defences}\label{def_eval}
\begin{figure}[t]
\center
\includegraphics[width=0.8\linewidth]{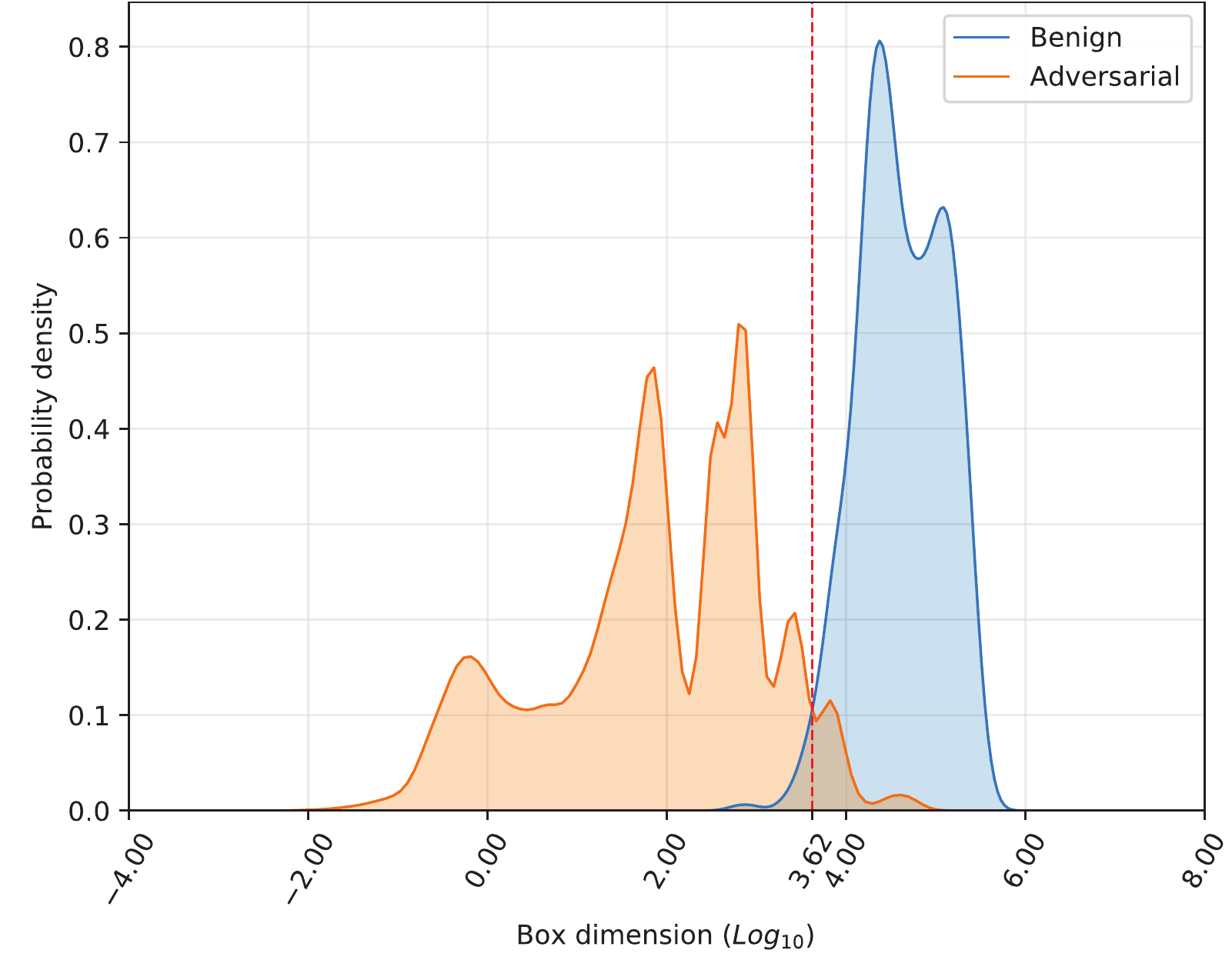}
\caption{Distributions of bounding box dimension from detection results of YOLO-v3. The distributions are based on detection results of 100 COCO images and corresponding 100 adversarial examples, respectively. In total, there are 812 boxes in the benign results and 286,469 adversarial boxes in the adversarial results. As a defence, boxes whose dimension is below $10^{3.62}$ in the NMS process are discarded.}
\label{box_dimension}
\end{figure}

\begin{figure}[t]
\center
\includegraphics[width=0.8\linewidth]{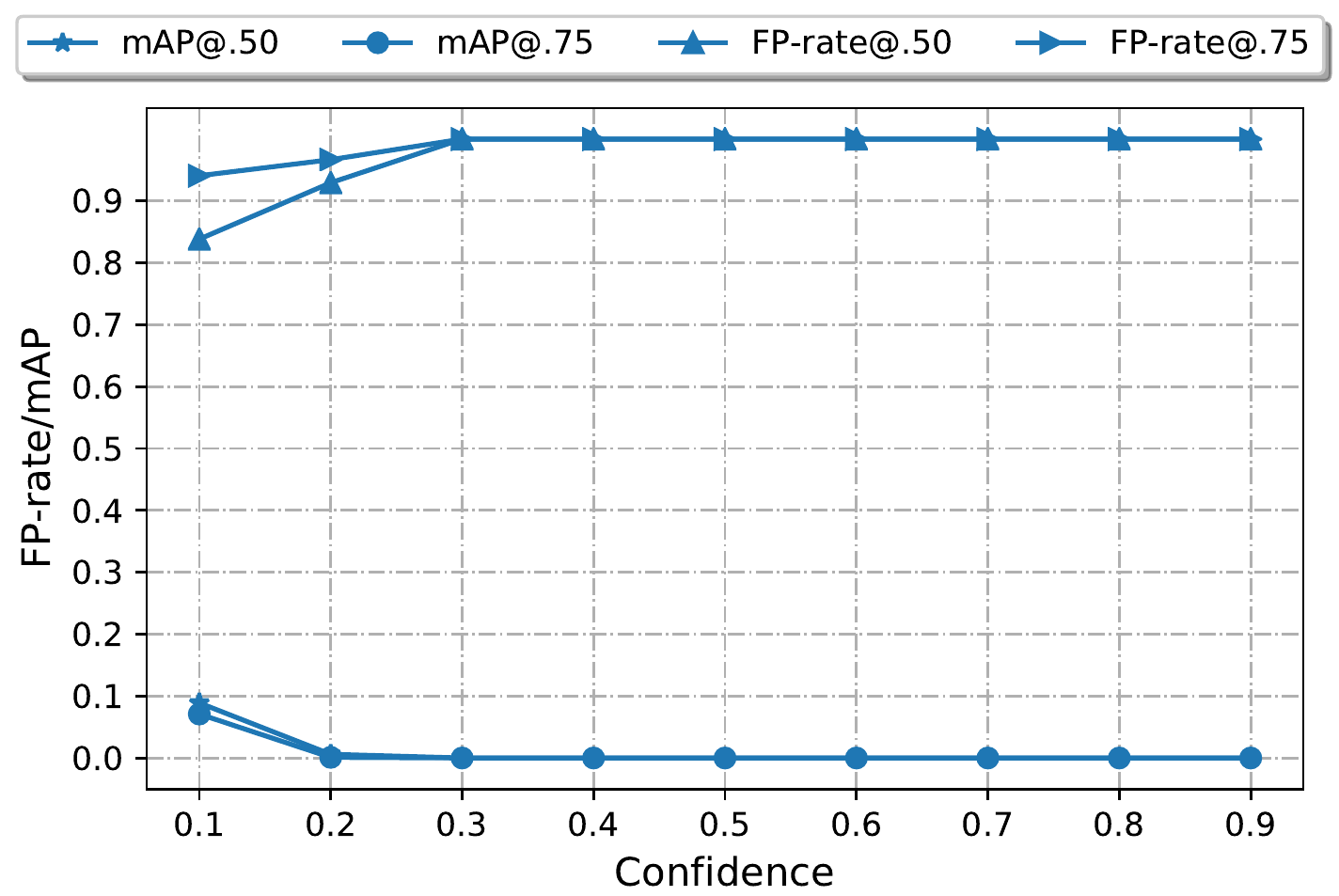}
\caption{The FP-rate and the mAP of YOLO-v3 with \textit{NMS defence}. The defence barely changes the FP-rate and the mAP. }
\label{NMS_defence_eval}
\end{figure}

Daedalus minimises detection boxes to trigger a NMS malfunction. Henceforth, a straight forward defence might be limiting the minimal allowed detection box size in the NMS process (\textit{i.e.}, \textit{NMS defence}). To investigate this possible defence, we first study the outputs of YOLO-v3 given 100 benign examples and corresponding 100 Daedalus examples. The distribution of the detection box dimension is visualised in Fig.~\ref{box_dimension}. According to Fig.~\ref{box_dimension}, an empirical threshold of $10^{3.62}$ is selected to filter adversarial boxes from detected boxes. Given this defence, the FP rate and the mAP of YOLO-v3 are again evaluated on $10$ Daedalus examples. The evaluation results are plotted in Fig.~\ref{NMS_defence_eval}. According to the results, \textit{NMS defence} is not an effective defence.

\begin{figure}[t]
\center
\includegraphics[width=0.8\linewidth]{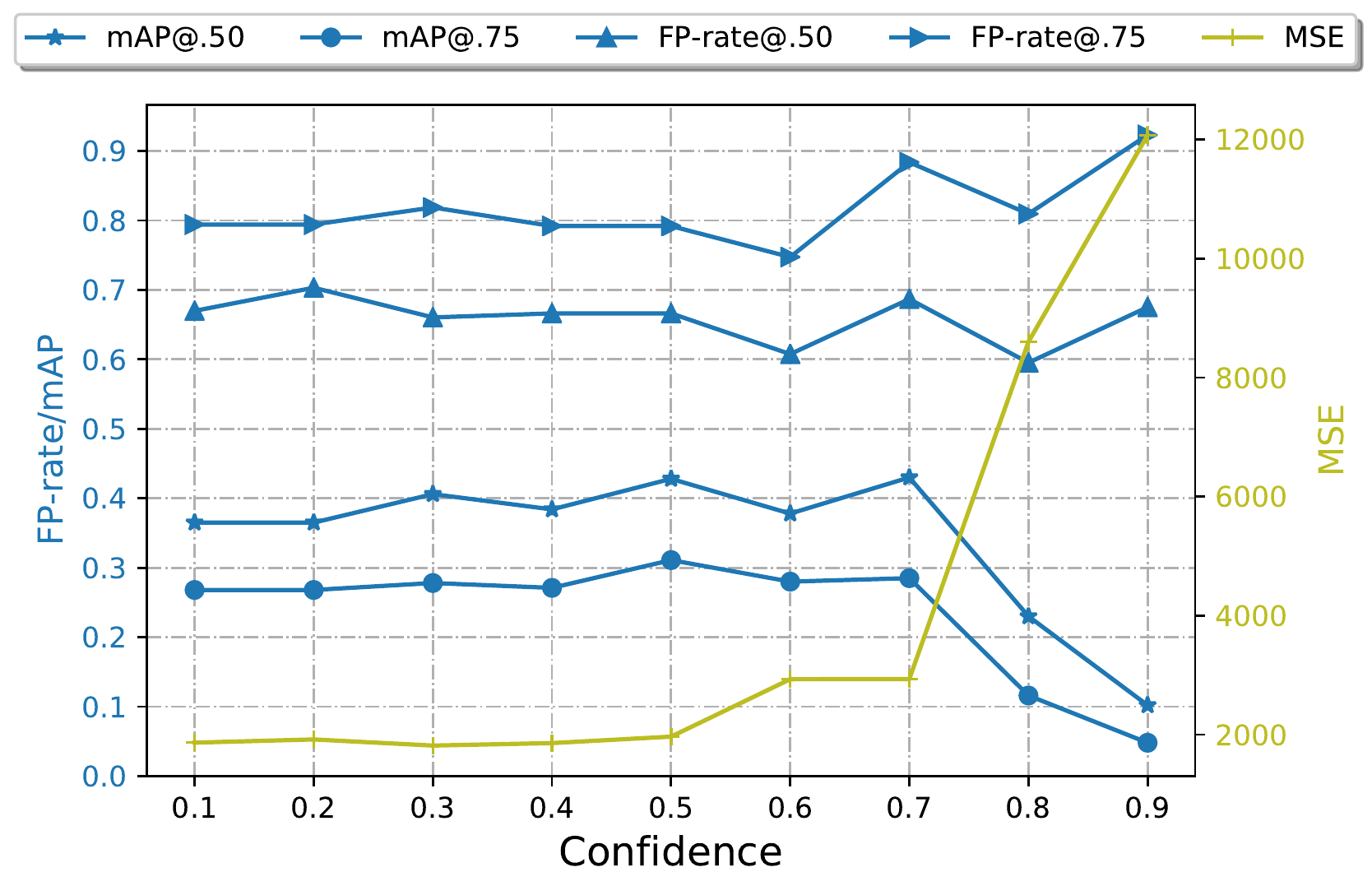}
\caption{The FP-rate and the mAP of YOLO-v3 with \textit{MagNet defence}. Reforming Daedalus examples by an autoencoder can slightly increase the performance of YOLO-v3. However, the FP-rate@.50 still stays above 0.6 while the FP-rate@.75 stays above 0.7. The mAP@.50 remains below 0.45 while the mAP@.75 remains below 0.3. Furthermore, the model performance drops when the confidence of the attack goes above 0.7.}
\label{MagNet_defence_eval}
\end{figure}

A second adaptive defence might be MagNet~\cite{meng2017magnet}. An denoise autoencoder is trained to reform Daedalus examples back to benign ones, and measure the reconstruction error to detect strong perturbations. Therefore, the evaluation of the defence is twofold: 1) whether the defence can increase mAP and reduce FP rate via reforming an adversarial example; 2) whether the reconstruction error can be detected when an adversarial example cannot be turned into a benign one. To answer these two questions, we measure the FP rate, the mAP and the Mean Squared Error (MSE) between a Daedalus example and a reformed example (\textit{i.e.}, the reconstruction error) of a defended YOLO-v3 model. The results are plotted in Fig.~\ref{MagNet_defence_eval}. According to the evaluation results, first, reforming the adversarial examples can barely increase the performance of the detection model. Second, The reconstruction error increases only after the confidence of the attack reaches $0.5$. This means it is difficult to detect attacks with confidence levels below $0.5$. Interestingly, it is observed that the detected objects in an image reformed by a denoise autoencoder cannot retain proper positions and categories. The Daedalus attack poses an intimidating threat to OD tasks. As revealed by the attack, NMS is no longer a secure solution for OD.

\section{Related Work}\label{Related_work}
Researchers have extensively studied the methods for crafting adversarial examples against machine-learning-based classifiers. The algorithms for generating such examples can be either gradient-based attacks or gradient-free attacks. The former type of attacks finds adversarial examples by minimising the cost of an adversarial objective set by the attacker, based on gradients information. The attacks can be either one-step gradient descent~\cite{Goodfellow2014Explaining} or iterative gradient descent~\cite{Szegedy2013Intriguing, kurakin2016adversarial, carlini2017}. Additionally, attackers can compute a Jacobian saliency map of features and perturb those salient features~\cite{Papernot2016The}. Based on these algorithms, there are evolved attacks that use different distortion metrics to make adversarial examples more imperceptible to human eyes~\cite{hosseini2018semantic,engstrom2017rotation,yang2018adversarial}. Recently, Expectation over Transformation (EoT) algorithm has been proposed to synthesise robust adversarial examples~\cite{Athalye2017, brown2017adversarial}. On the other hand, gradient-free attacks rely on zeroth-order optimisation techniques to search adversarial examples against black-box models~\cite{chen2017zoo, liu2018zeroth, bhagoji2018practical, alzantot2018genattack}.

Beyond the above attacks, there are attacks targeting at domains such as malware detection and object detection~\cite{chen2019android}. For example, dense adversarial generation (DAG) algorithm was proposed to create adversarial examples for OD and segmentation~\cite{xie2017adversarial}. Lu et al. crafted robust adversarial examples that caused misclassification in object detector~\cite{lu2017adversarial}. $RP_2$ was proposed to craft adversarial example of real-world road signs~\cite{evtimov2017robust}. Subsequently, $RP_2$ was extended to attack YOLO-v2~\cite{song2018physical}. There are also other attempts to craft more robust physical examples~\cite{chen2018shapeshifter,thys2019fooling,zhao2019seeing}. However, current attacks mainly result in misclassification/appearance/disappearance of the detected objects~\cite{chen2018shapeshifter,thys2019fooling}. The Daedalus attack creates adversarial examples that cause malfunctioning of NMS, which is different from all the previous attacks in nature. 

\section{Conclusion}\label{Conclusion}
In this paper, we propose a novel type of adversarial example which aims at disabling NMS functionality in object detection models. The proposed attack, named Daedalus, can control both the strength of the generated adversarial examples and the object class to be attacked. The attacked model will output extremely noisy detection results such that it is no longer functional. The attack can reduce the mAP of detection to nearly $0\%$. The false positive rate for Daedalus examples can go up to $99.9\%$. Meanwhile, the required distortion to launch the Daedalus attack is imperceptible. This attack aims at breaking NMS instead of causing misclassifications. Unlike misclassification-targeting adversarial examples, it is difficult to defend against such attack since the attacked feature map is much more complicated than the classification logits. Transferability of the Daedalus attack is affected by the selected substitutes. We rely on minimising the expectation of box dimensions over a set of feature maps to make robust examples, which can make NMS malfunction in multiple detection models. There are some remaining problems we aim to address in future. For example, considering the complexity of OD feature maps, it is difficult to make universal adversarial examples that can launch zero-knowledge attacks. The Daedalus attack reveals a vulnerability lies in object detection algorithms, which can have fatal consequences in the real world. Nevertheless, defending method against Daedalus is still a missing piece. We will also try to address this problem in our future work.

\bibliographystyle{IEEEtran}
\bibliography{../advref}

\begin{thebibliography}{10}
\providecommand{\url}[1]{#1}
\csname url@samestyle\endcsname
\providecommand{\newblock}{\relax}
\providecommand{\bibinfo}[2]{#2}
\providecommand{\BIBentrySTDinterwordspacing}{\spaceskip=0pt\relax}
\providecommand{\BIBentryALTinterwordstretchfactor}{4}
\providecommand{\BIBentryALTinterwordspacing}{\spaceskip=\fontdimen2\font plus
\BIBentryALTinterwordstretchfactor\fontdimen3\font minus
  \fontdimen4\font\relax}
\providecommand{\BIBforeignlanguage}[2]{{%
\expandafter\ifx\csname l@#1\endcsname\relax
\typeout{** WARNING: IEEEtran.bst: No hyphenation pattern has been}%
\typeout{** loaded for the language `#1'. Using the pattern for}%
\typeout{** the default language instead.}%
\else
\language=\csname l@#1\endcsname
\fi
#2}}
\providecommand{\BIBdecl}{\relax}
\BIBdecl

\bibitem{lin2014microsoft}
T.-Y. Lin, M.~Maire, S.~Belongie, J.~Hays, P.~Perona, D.~Ramanan,
  P.~Doll{\'a}r, and C.~L. Zitnick, ``Microsoft coco: Common objects in
  context,'' in \emph{Proceedings of the European Conference on Computer Vision
  (ECCV)}.\hskip 1em plus 0.5em minus 0.4em\relax Springer, 2014, pp. 740--755.

\bibitem{liu2016ssd}
W.~Liu, D.~Anguelov, D.~Erhan, C.~Szegedy, S.~Reed, C.-Y. Fu, and A.~C. Berg,
  ``Ssd: Single shot multibox detector,'' in \emph{Proceedings of the European
  Conference on Computer Vision (ECCV)}.\hskip 1em plus 0.5em minus 0.4em\relax
  Springer, 2016, pp. 21--37.

\bibitem{lin2017focal}
T.-Y. Lin, P.~Goyal, R.~Girshick, K.~He, and P.~Doll{\'a}r, ``Focal loss for
  dense object detection,'' in \emph{Proceedings of the IEEE International
  Conference on Computer Vision (ICCV)}.\hskip 1em plus 0.5em minus 0.4em\relax
  IEEE, 2017, pp. 2980--2988.

\bibitem{redmon2016you}
J.~Redmon, S.~Divvala, R.~Girshick, and A.~Farhadi, ``You only look once:
  Unified, real-time object detection,'' in \emph{Proceedings of the IEEE
  conference on Computer Vision and Pattern Recognition (CVPR)}.\hskip 1em plus
  0.5em minus 0.4em\relax IEEE, 2016, pp. 779--788.

\bibitem{girshick2014rich}
R.~Girshick, J.~Donahue, T.~Darrell, and J.~Malik, ``Rich feature hierarchies
  for accurate object detection and semantic segmentation,'' in
  \emph{Proceedings of the IEEE conference on Computer Vision and Pattern
  Recognition (CVPR)}.\hskip 1em plus 0.5em minus 0.4em\relax IEEE, 2014, pp.
  580--587.

\bibitem{girshick2015fast}
R.~Girshick, ``Fast r-cnn,'' in \emph{Proceedings of the IEEE International
  Vonference on Computer Vision (ICCV)}.\hskip 1em plus 0.5em minus 0.4em\relax
  IEEE, 2015, pp. 1440--1448.

\bibitem{neubeck2006efficient}
A.~Neubeck and L.~Van~Gool, ``Efficient non-maximum suppression,'' in
  \emph{Proceedings of the International Conference on Pattern Recognition
  (ICPR)}, vol.~3.\hskip 1em plus 0.5em minus 0.4em\relax IEEE, 2006, pp.
  850--855.

\bibitem{cong2018speedup}
Y.~Cong, D.~Tian, Y.~Feng, B.~Fan, and H.~Yu, ``Speedup 3-d texture-less object
  recognition against self-occlusion for intelligent manufacturing,''
  \emph{IEEE transactions on cybernetics}, vol.~49, no.~11, pp. 3887--3897,
  2018.

\bibitem{Schroff2015FaceNet}
F.~Schroff, D.~Kalenichenko, and J.~Philbin, ``Facenet: A unified embedding for
  face recognition and clustering,'' in \emph{Proceedings of the IEEE
  conference on Computer Vision and Pattern Recognition (CVPR)}.\hskip 1em plus
  0.5em minus 0.4em\relax IEEE, 2015, pp. 815--823.

\bibitem{wu2018face}
W.~Wu, Y.~Yin, X.~Wang, and D.~Xu, ``Face detection with different scales based
  on faster r-cnn,'' \emph{IEEE transactions on cybernetics}, vol.~49, no.~11,
  pp. 4017--4028, 2018.

\bibitem{hosang2017learning}
J.~H. Hosang, R.~Benenson, and B.~Schiele, ``Learning non-maximum
  suppression.'' in \emph{Proceedings of the IEEE conference on Computer Vision
  and Pattern Recognition (CVPR)}.\hskip 1em plus 0.5em minus 0.4em\relax IEEE,
  2017, pp. 6469--6477.

\bibitem{bodla2017soft}
N.~Bodla, B.~Singh, R.~Chellappa, and L.~S. Davis, ``Soft-nms-improving object
  detection with one line of code,'' in \emph{Proceedings of the IEEE
  International Conference on Computer Vision (ICCV)}.\hskip 1em plus 0.5em
  minus 0.4em\relax IEEE, 2017, pp. 5562--5570.

\bibitem{Akhtar2018Threat}
N.~Akhtar and A.~Mian, ``Threat of adversarial attacks on deep learning in
  computer vision: A survey,'' \emph{IEEE Access}, vol.~6, pp.
  14\,410--14\,430, 2018.

\bibitem{lu2017adversarial}
J.~Lu, H.~Sibai, and E.~Fabry, ``Adversarial examples that fool detectors,''
  \emph{arXiv preprint arXiv:1712.02494}, 2017.

\bibitem{xie2017adversarial}
C.~Xie, J.~Wang, Z.~Zhang, Y.~Zhou, L.~Xie, and A.~Yuille, ``Adversarial
  examples for semantic segmentation and object detection,'' in
  \emph{Proceedings of the IEEE International Vonference on Computer Vision
  (ICCV)}.\hskip 1em plus 0.5em minus 0.4em\relax IEEE, 2017, pp. 1369--1378.

\bibitem{song2018physical}
D.~Song, K.~Eykholt, I.~Evtimov, E.~Fernandes, B.~Li, A.~Rahmati, F.~Tramer,
  A.~Prakash, and T.~Kohno, ``Physical adversarial examples for object
  detectors,'' in \emph{Proceedings of the 12th USENIX Workshop on Offensive
  Technologies (WOOT 18)}.\hskip 1em plus 0.5em minus 0.4em\relax USENIX
  Association, 2018.

\bibitem{chen2018shapeshifter}
S.-T. Chen, C.~Cornelius, J.~Martin, and D.~H.~P. Chau, ``Shapeshifter: Robust
  physical adversarial attack on faster r-cnn object detector,'' in \emph{Joint
  European Conference on Machine Learning and Knowledge Discovery in
  Databases}.\hskip 1em plus 0.5em minus 0.4em\relax Springer, 2018, pp.
  52--68.

\bibitem{thys2019fooling}
S.~Thys, W.~Van~Ranst, and T.~Goedem{\'e}, ``Fooling automated surveillance
  cameras: adversarial patches to attack person detection,'' in
  \emph{Proceedings of the IEEE conference on Computer Vision and Pattern
  Recognition (CVPR) Workshops}.\hskip 1em plus 0.5em minus 0.4em\relax IEEE,
  2019, pp. 0--0.

\bibitem{zhao2019seeing}
Y.~Zhao, H.~Zhu, R.~Liang, Q.~Shen, S.~Zhang, and K.~Chen, ``Seeing isn't
  believing: Towards more robust adversarial attack against real world object
  detectors,'' in \emph{Proceedings of the 2019 ACM SIGSAC Conference on
  Computer and Communications Security}.\hskip 1em plus 0.5em minus 0.4em\relax
  ACM, 2019, pp. 1989--2004.

\bibitem{verma2016object}
N.~K. Verma, T.~Sharma, S.~D. Rajurkar, and A.~Salour, ``Object identification
  for inventory management using convolutional neural network,'' in \emph{2016
  IEEE Applied Imagery Pattern Recognition Workshop (AIPR)}.\hskip 1em plus
  0.5em minus 0.4em\relax IEEE, 2016, pp. 1--6.

\bibitem{yang2018deep}
Z.~Yang, W.~Yu, P.~Liang, H.~Guo, L.~Xia, F.~Zhang, Y.~Ma, and J.~Ma, ``Deep
  transfer learning for military object recognition under small training set
  condition,'' \emph{Neural Computing and Applications}, pp. 1--10, 2018.

\bibitem{redmon2018yolov3}
J.~Redmon and A.~Farhadi, ``Yolov3: An incremental improvement,'' \emph{arXiv
  preprint arXiv:1804.02767}, 2018.

\bibitem{liu2018deep}
L.~Liu, W.~Ouyang, X.~Wang, P.~Fieguth, J.~Chen, X.~Liu, and
  M.~Pietik{\"a}inen, ``Deep learning for generic object detection: A survey,''
  \emph{arXiv preprint arXiv:1809.02165}, 2018.

\bibitem{everingham2010pascal}
M.~Everingham, L.~Van~Gool, C.~K. Williams, J.~Winn, and A.~Zisserman, ``The
  pascal visual object classes (voc) challenge,'' \emph{International Journal
  of Computer Vision}, vol.~88, no.~2, pp. 303--338, 2010.

\bibitem{shafiee2017fast}
M.~J. Shafiee, B.~Chywl, F.~Li, and A.~Wong, ``Fast yolo: A fast you only look
  once system for real-time embedded object detection in video,'' \emph{arXiv
  preprint arXiv:1709.05943}, 2017.

\bibitem{lu2017efficient}
K.~Lu, X.~An, J.~Li, and H.~He, ``Efficient deep network for vision-based
  object detection in robotic applications,'' \emph{Neurocomputing}, vol. 245,
  pp. 31--45, 2017.

\bibitem{zhang2017real}
J.~Zhang, M.~Huang, X.~Jin, and X.~Li, ``A real-time chinese traffic sign
  detection algorithm based on modified yolov2,'' \emph{Algorithms}, vol.~10,
  no.~4, p. 127, 2017.

\bibitem{carlini2017}
N.~Carlini and D.~Wagner, ``Towards evaluating the robustness of neural
  networks,'' in \emph{Proceedings of the IEEE Symposium on Security and
  Privacy (S\&P)}.\hskip 1em plus 0.5em minus 0.4em\relax IEEE, 2017, pp.
  39--57.

\bibitem{gurbaxani2018traits}
R.~Gurbaxani and S.~Mishra, ``Traits \& transferability of adversarial examples
  against instance segmentation \& object detection,'' \emph{arXiv preprint
  arXiv:1808.01452}, 2018.

\bibitem{he2016deep}
K.~He, X.~Zhang, S.~Ren, and J.~Sun, ``Deep residual learning for image
  recognition,'' in \emph{Proceedings of the IEEE conference on Computer Vision
  and Pattern Recognition (CVPR)}.\hskip 1em plus 0.5em minus 0.4em\relax IEEE,
  2016, pp. 770--778.

\bibitem{sener2018multi}
O.~Sener and V.~Koltun, ``Multi-task learning as multi-objective
  optimization,'' in \emph{Advances in Neural Information Processing Systems},
  2018, pp. 527--538.

\bibitem{sharif2016accessorize}
M.~Sharif, S.~Bhagavatula, L.~Bauer, and M.~K. Reiter, ``Accessorize to a
  crime: Real and stealthy attacks on state-of-the-art face recognition,'' in
  \emph{Proceedings of the ACM SIGSAC Conference on Computer and Communications
  Security}.\hskip 1em plus 0.5em minus 0.4em\relax ACM, 2016, pp. 1528--1540.

\bibitem{meng2017magnet}
D.~Meng and H.~Chen, ``Magnet: a two-pronged defense against adversarial
  examples,'' \emph{arXiv preprint arXiv:1705.09064}, 2017.

\bibitem{Goodfellow2014Explaining}
I.~J. Goodfellow, J.~Shlens, and C.~Szegedy, ``Explaining and harnessing
  adversarial examples,'' \emph{Computer Science}, 2014.

\bibitem{Szegedy2013Intriguing}
C.~Szegedy, W.~Zaremba, I.~Sutskever, J.~Bruna, D.~Erhan, I.~Goodfellow, and
  R.~Fergus, ``Intriguing properties of neural networks,'' \emph{Computer
  Science}, 2013.

\bibitem{kurakin2016adversarial}
A.~Kurakin, I.~Goodfellow, and S.~Bengio, ``Adversarial examples in the
  physical world,'' \emph{arXiv preprint arXiv:1607.02533}, 2016.

\bibitem{Papernot2016The}
N.~Papernot, P.~Mcdaniel, S.~Jha, M.~Fredrikson, Z.~B. Celik, and A.~Swami,
  ``The limitations of deep learning in adversarial settings,'' pp. 372--387,
  2016.

\bibitem{hosseini2018semantic}
H.~Hosseini and R.~Poovendran, ``Semantic adversarial examples,'' in
  \emph{Proceedings of the IEEE conference on Computer Vision and Pattern
  Recognition (CVPR) Workshops}.\hskip 1em plus 0.5em minus 0.4em\relax IEEE,
  2018, pp. 1614--1619.

\bibitem{engstrom2017rotation}
L.~Engstrom, D.~Tsipras, L.~Schmidt, and A.~Madry, ``A rotation and a
  translation suffice: Fooling cnns with simple transformations,'' \emph{arXiv
  preprint arXiv:1712.02779}, 2017.

\bibitem{yang2018adversarial}
E.~Yang, T.~Liu, C.~Deng, and D.~Tao, ``Adversarial examples for hamming space
  search,'' \emph{IEEE transactions on cybernetics}, vol.~PP, no.~99, pp.
  1--12, 2018.

\bibitem{Athalye2017}
A.~Athalye and I.~Sutskever, ``Synthesizing robust adversarial examples,''
  \emph{arXiv preprint arXiv:1707.07397}, 2017.

\bibitem{brown2017adversarial}
T.~B. Brown, D.~Man{\'e}, A.~Roy, M.~Abadi, and J.~Gilmer, ``Adversarial
  patch,'' \emph{arXiv preprint arXiv:1712.09665}, 2017.

\bibitem{chen2017zoo}
P.-Y. Chen, H.~Zhang, Y.~Sharma, J.~Yi, and C.-J. Hsieh, ``Zoo: Zeroth order
  optimization based black-box attacks to deep neural networks without training
  substitute models,'' in \emph{Proceedings of the 10th ACM Workshop on
  Artificial Intelligence and Security}.\hskip 1em plus 0.5em minus 0.4em\relax
  ACM, 2017, pp. 15--26.

\bibitem{liu2018zeroth}
S.~Liu, B.~Kailkhura, P.-Y. Chen, P.~Ting, S.~Chang, and L.~Amini,
  ``Zeroth-order stochastic variance reduction for nonconvex optimization,'' in
  \emph{Advances in Neural Information Processing Systems}, 2018, pp.
  3727--3737.

\bibitem{bhagoji2018practical}
A.~N. Bhagoji, W.~He, B.~Li, and D.~Song, ``Practical black-box attacks on deep
  neural networks using efficient query mechanisms,'' in \emph{Proceedings of
  the European Conference on Computer Vision (ECCV)}.\hskip 1em plus 0.5em
  minus 0.4em\relax Springer, 2018, pp. 158--174.

\bibitem{alzantot2018genattack}
M.~Alzantot, Y.~Sharma, S.~Chakraborty, and M.~Srivastava, ``Genattack:
  Practical black-box attacks with gradient-free optimization,'' \emph{arXiv
  preprint arXiv:1805.11090}, 2018.

\bibitem{chen2019android}
X.~Chen, C.~Li, D.~Wang, S.~Wen, J.~Zhang, S.~Nepal, Y.~Xiang, and K.~Ren,
  ``Android hiv: A study of repackaging malware for evading machine-learning
  detection,'' \emph{IEEE Transactions on Information Forensics and Security},
  vol.~15, pp. 987--1001, 2019.

\bibitem{evtimov2017robust}
I.~Evtimov, K.~Eykholt, E.~Fernandes, T.~Kohno, B.~Li, A.~Prakash, A.~Rahmati,
  and D.~Song, ``Robust physical-world attacks on machine learning models,''
  \emph{arXiv preprint arXiv:1707.08945}, vol.~2, no.~3, p.~4, 2017.

\end{thebibliography}

\appendix
\section{Digital attacks}
\IEEEPARstart{M}{ore} results of the digital Daedalus attack are included in this section.

\begin{figure}[h!]
\center
\includegraphics[width=1\linewidth]{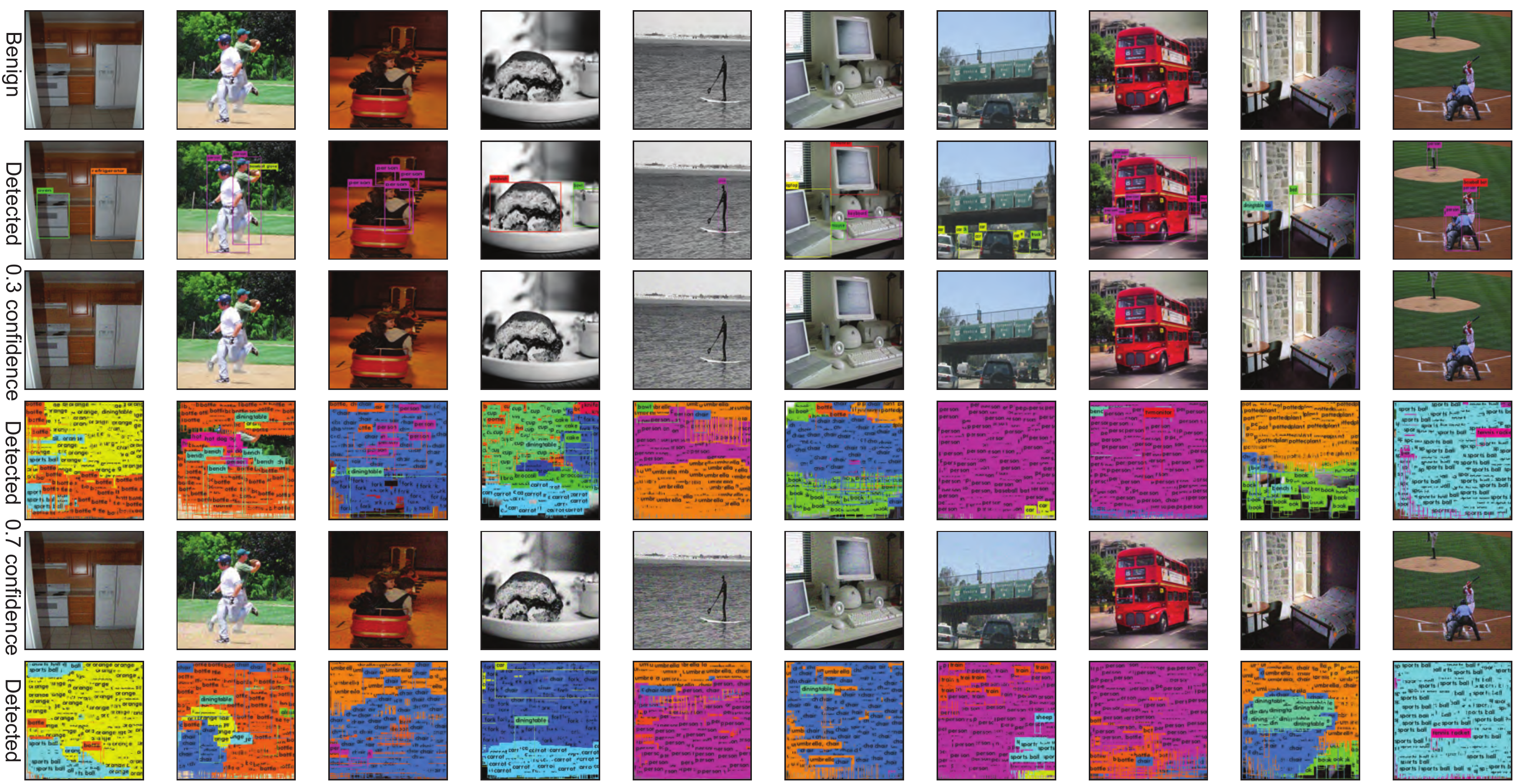}
\caption{Adversarial examples made by the $L_0$ attack. The first row contains the original images. The third row contains the low-confidence ($\gamma=0.3$) adversarial examples. The fifth row contains the high-confidence ($\gamma=0.7$) examples. The detection results from YOLO-v3 are in the rows below them. The confidence controls the density of the redundant detection boxes in the detection results.}
\label{l0_demo}
\end{figure}

\begin{figure}[h!]
\center
\includegraphics[width=1\linewidth]{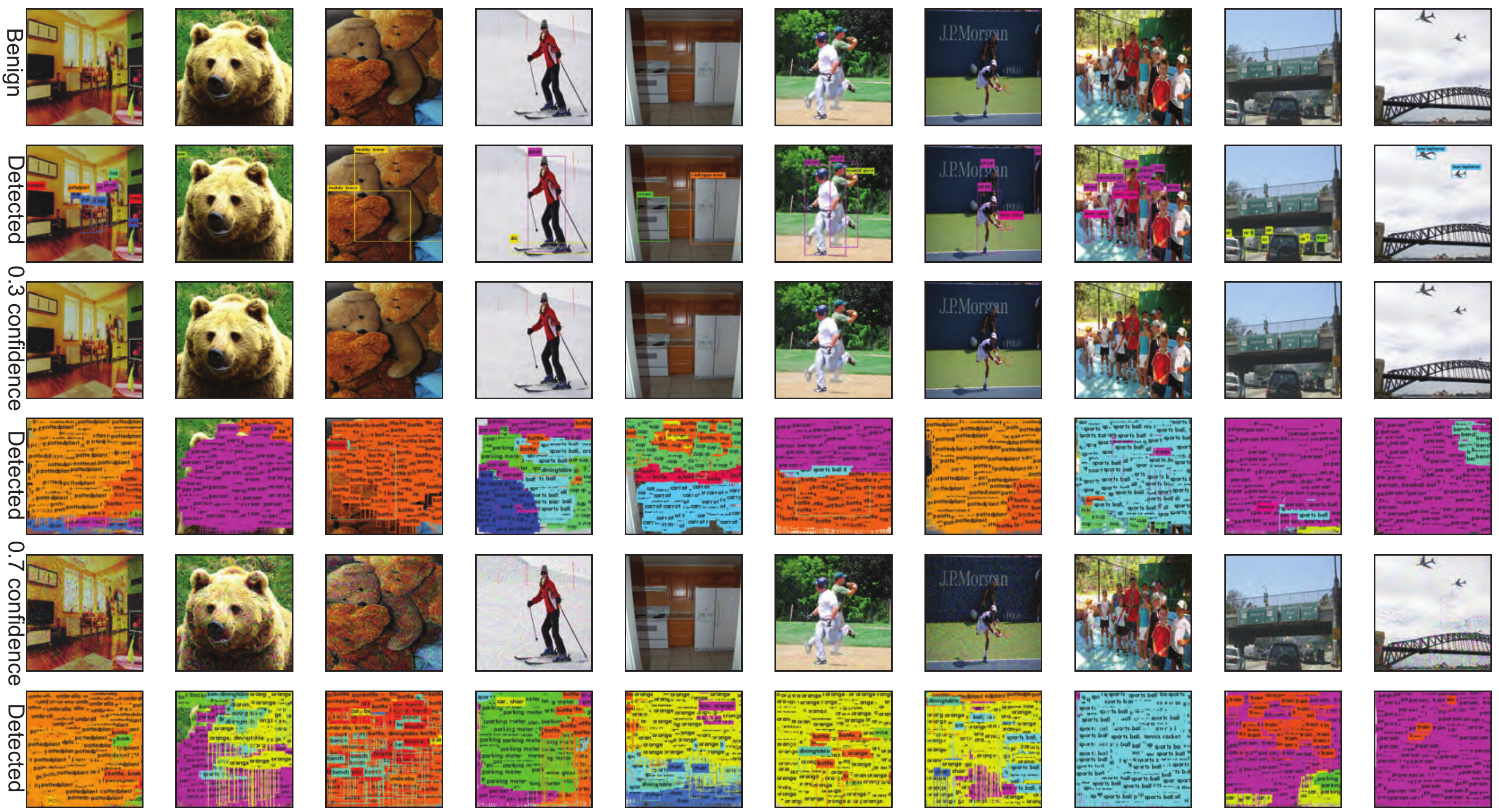}
\caption{Adversarial examples made by the $L_2$ attack. The first row contains the original images. The third row contains the low-confidence ($\gamma=0.3$) adversarial examples. The fifth row contains the high-confidence ($\gamma=0.7$) examples. The detection results from YOLO-v3 are in the rows below them. The confidence controls the density of the redundant detection boxes in the detection results.}
\label{l2_demo}
\end{figure}

\begin{figure}[h!]
\center
\includegraphics[width=1\linewidth]{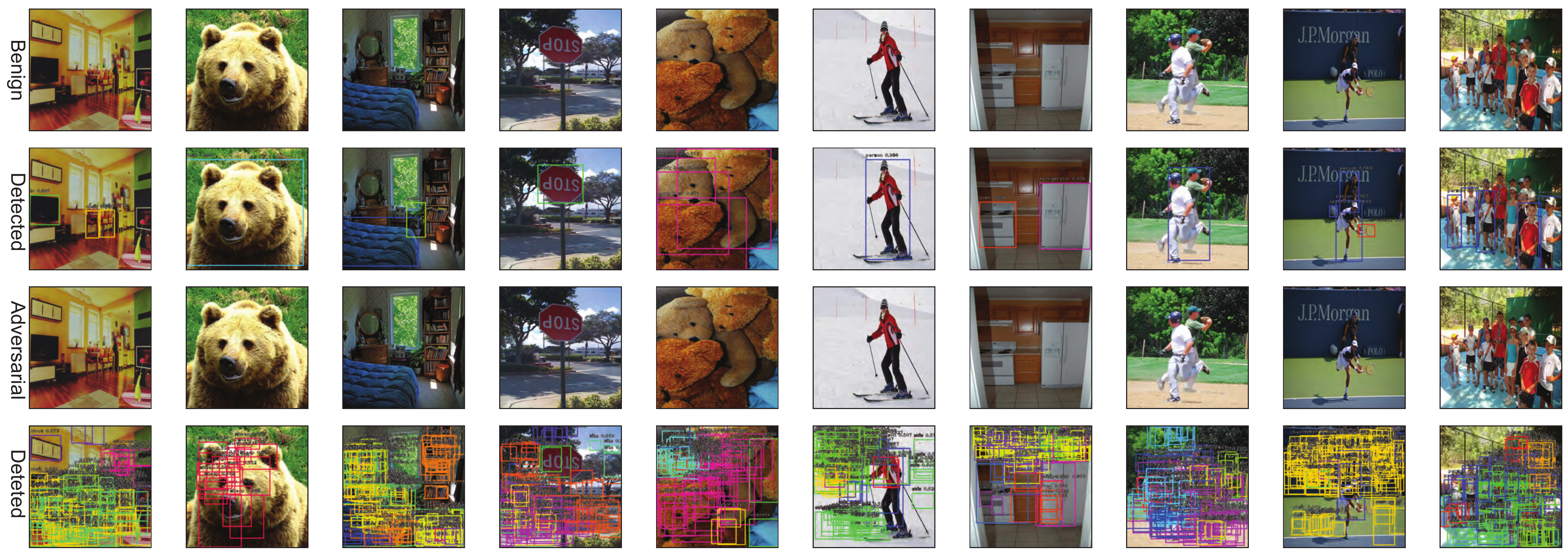}
\caption{The detection results of Daedalus adversarial examples made by the $L_2$ attack towards RetinaNet-ResNet-50. The adversarial examples are crafted based on a confidence of $0.3$.}
\label{l2_retinanet}
\end{figure}

\begin{figure}[h!]
\center
\includegraphics[width=1\linewidth]{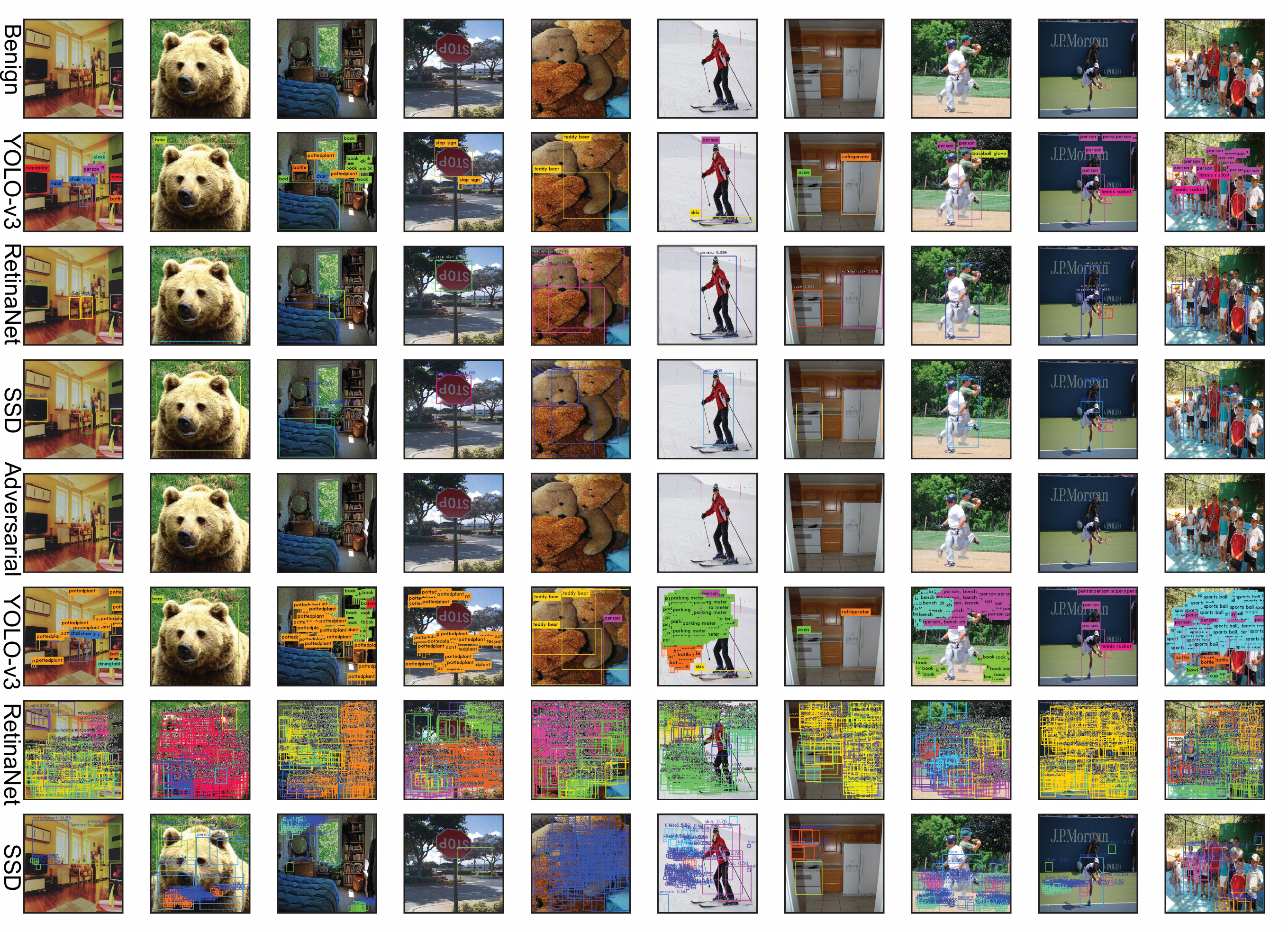}
\caption{The detection results of Daedalus adversarial examples made by the ensemble $L_2$ attack towards YOLO-v3, RetinaNet, and SSD. The adversarial examples are crafted based on a confidence of $0.3$. The first row displays the benign examples. The second, the third, and the fourth rows are the detection results of the benign examples from YOLO-v3, RetinaNet, and SSD, respectively. The Daedalus examples are displayed in the fourth row. The detection results of the adversarial examples from YOLO-v3, RetinaNet, and SSD are in the following rows.}
\label{l2_ensemble}
\end{figure}

\begin{figure}[h!]
\center
\includegraphics[width=1\linewidth]{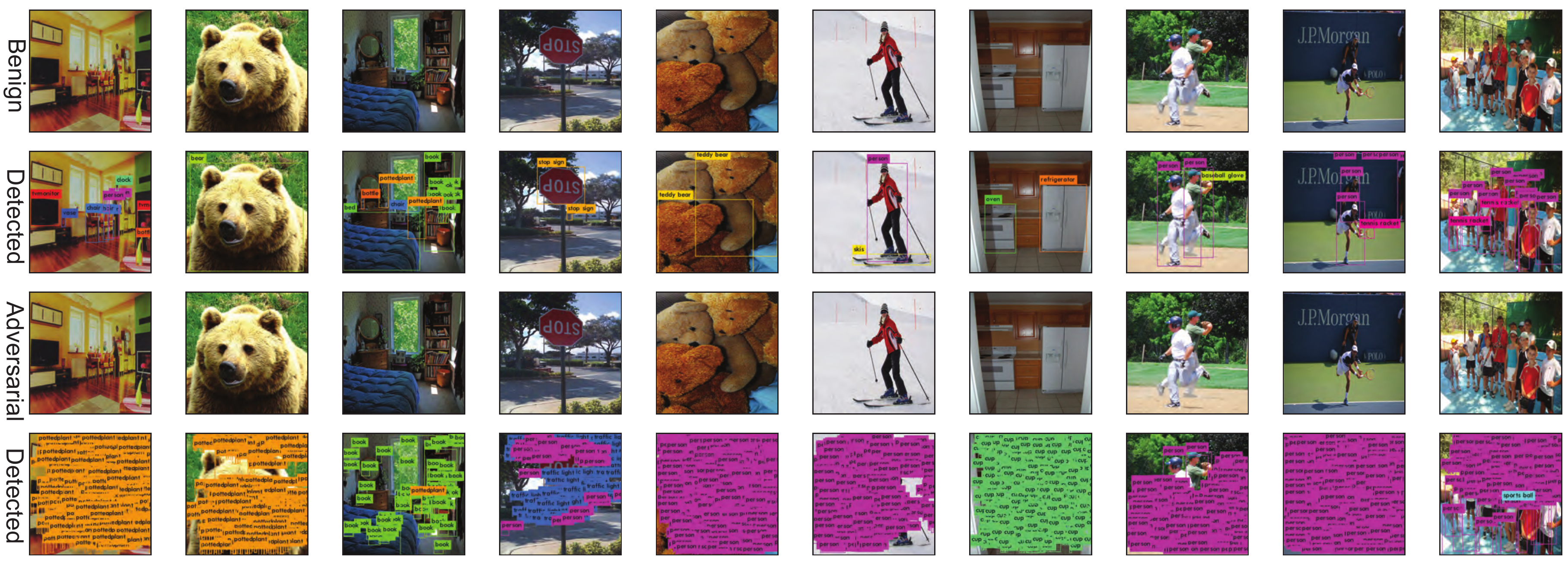}
\caption{Attacking the object category which has the most number of the detection boxes before NMS. The attacked model is YOLO-v3.}
\label{l2_mostclass}
\end{figure}

\begin{figure}[h!]
\center
\includegraphics[width=1\linewidth]{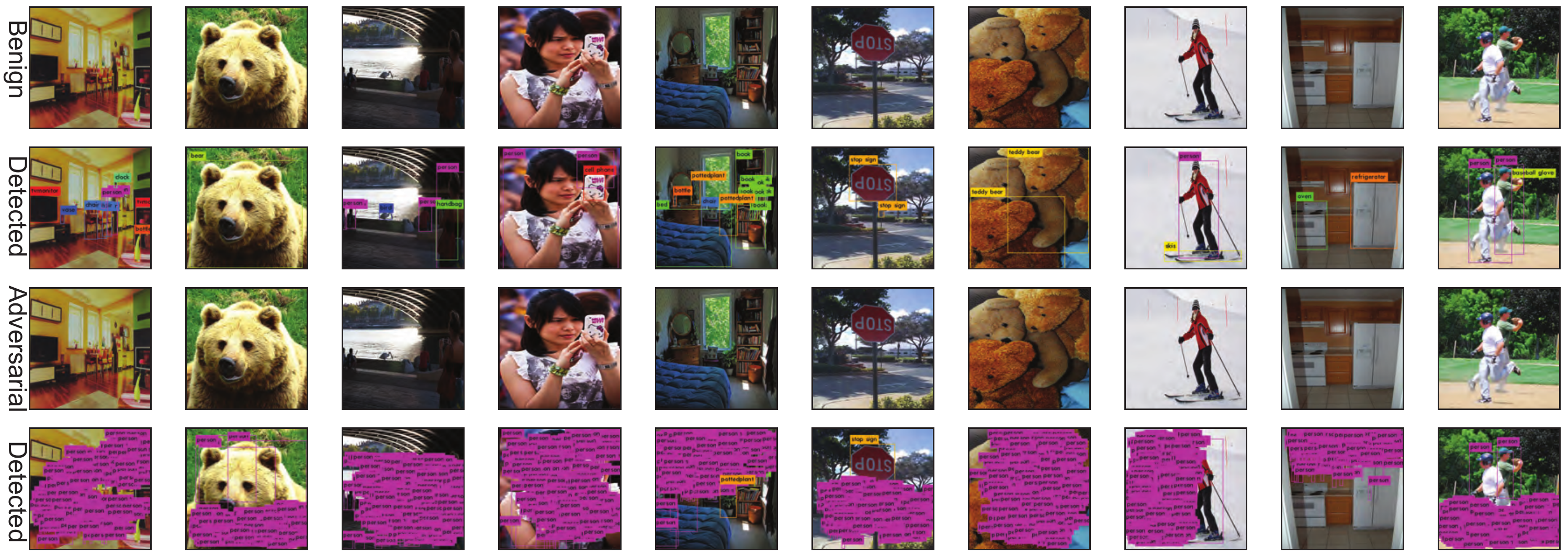}
\caption{Demonstration of attacking object in a specified category for YOLO-v3. We select `person' to attack.}
\label{l2_person}
\end{figure}

\begin{figure}[h!]
\center
\includegraphics[width=1\linewidth]{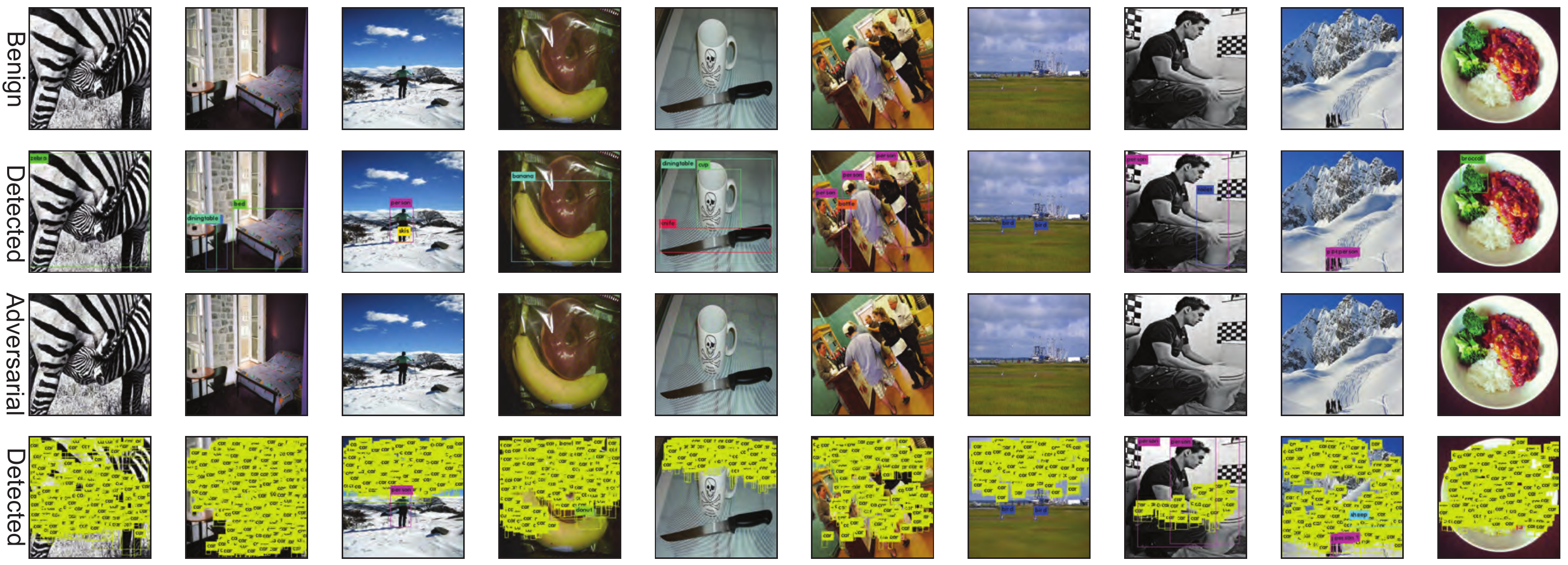}
\caption{Demonstration of attacking object in a specified category for YOLO-v3. We select `car' to attack.}
\label{l2_car}
\end{figure}

\FloatBarrier
\section{Physical attack}
\subsection{Parameters of the attack}
The details of the transformations used for crafting the Daedalus posters are summarised in Table~\ref{tran}. The table presents the details for making a $w_p\times h_p$ poster.

\begin{table*}[h!]
\caption{List of Transformations}
\label{tran}
\centering
\begin{tabular}{c|c}
\hline
Inputs & Perturbation size: $w_p\times h_p$, video frame size: $w_v\times h_v$\\
\hline
Transformation & Parameter\\
\hline
Random noise & $U(0,0.01)$ \\
Aspect ratio ($\alpha$) & $416/w_v:416/h_v$\\
Zooming ($z$) & $0.1\times \min(\frac{w_v}{w_p}, \frac{h_v}{h_p}) \sim 0.7\times \min(\frac{w_v}{w_p}, \frac{h_v}{h_p})$\\
Rotation & $-\pi/10\to \pi/10$\\
Position (top left coordinates) & $(x\sim U(0, 416-\alpha\cdot z\cdot w_p)$, $y\sim U(0, 416-\alpha\cdot z\cdot h_p)$\\
SNPS rate ($\beta$) & $0.001\sim 0.01$\\
\hline
\end{tabular}
\end{table*}

\subsection{Demonstration of the attack}
\begin{figure}[h!]
\centering
  \begin{subfigure}[b]{.45\linewidth}
    \centering
    \includegraphics[width=0.95\linewidth]{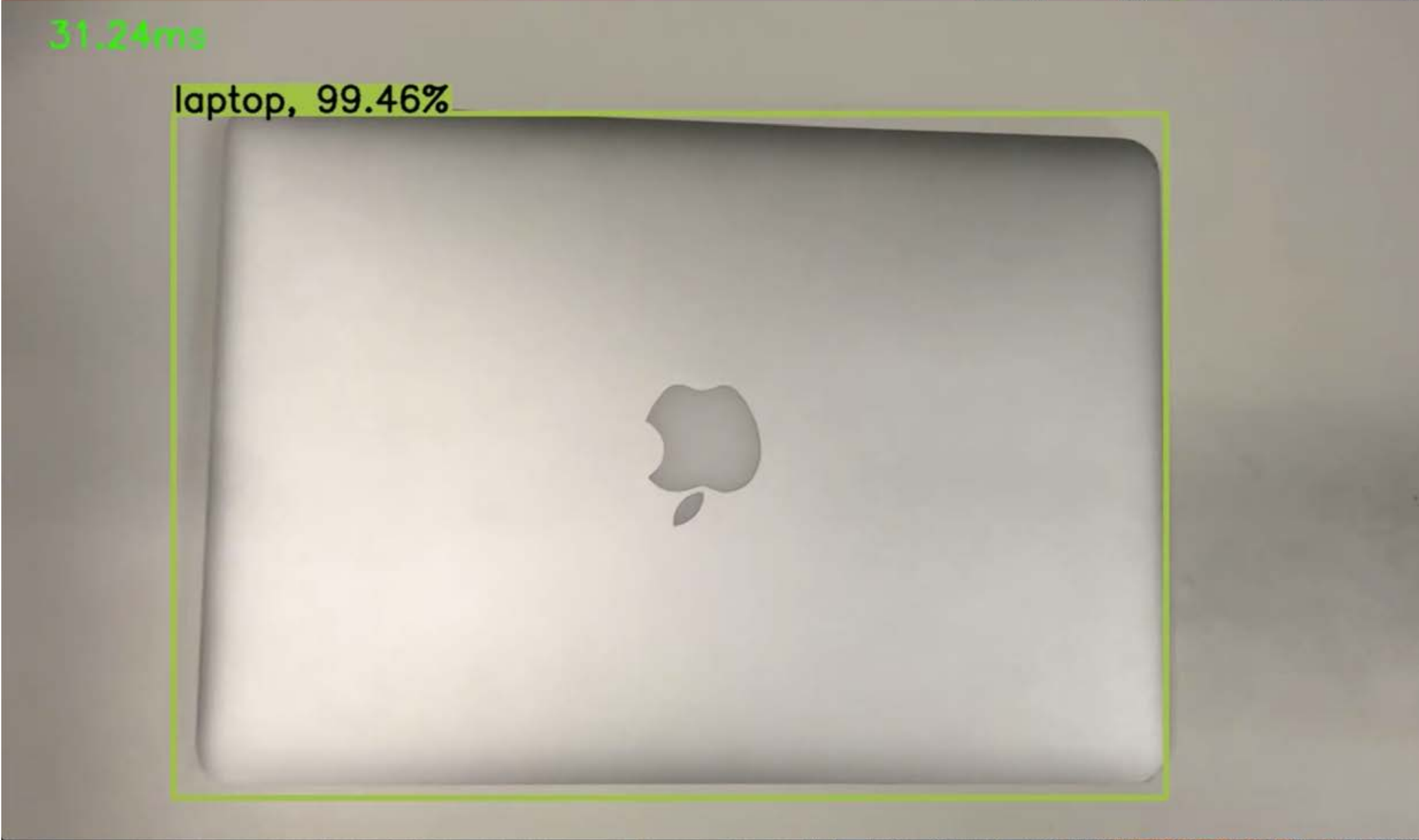}
	\caption{}
	\label{no_poster_mac}
  \end{subfigure}%
  \begin{subfigure}[b]{.45\linewidth}
    \centering
    \includegraphics[width=1\linewidth]{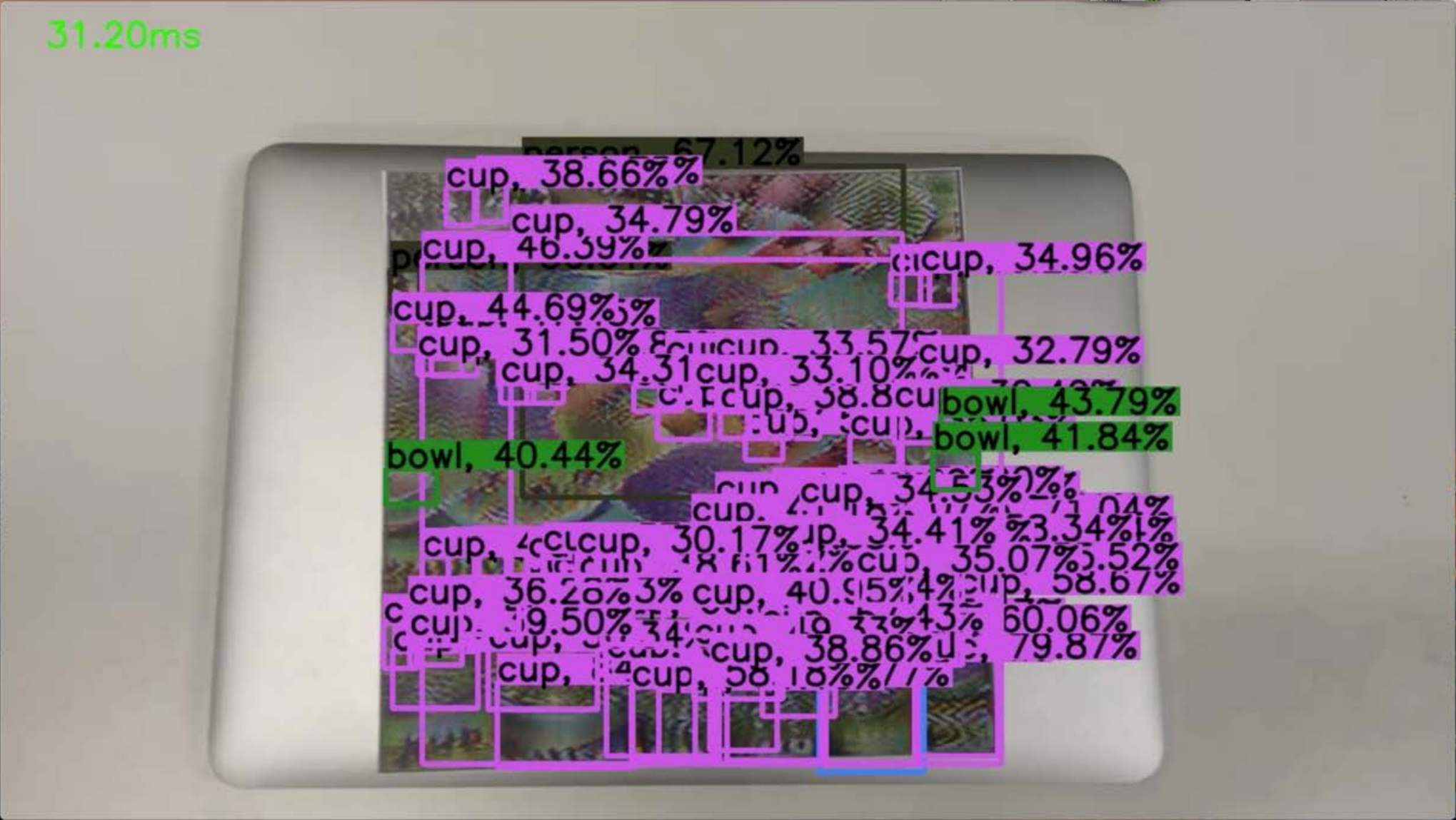}
    \caption{}
    \label{adv_mac}
  \end{subfigure}%
  \caption{(a) Detection results of a MacBook in a benign setting. (b) Detection results of the MacBook when we apply a Daedalus poster attacking all object categories.}
  \label{mac}
\end{figure}

\begin{figure}[h!]
\centering
  \begin{subfigure}[b]{.45\linewidth}
    \centering
    \includegraphics[width=1\linewidth]{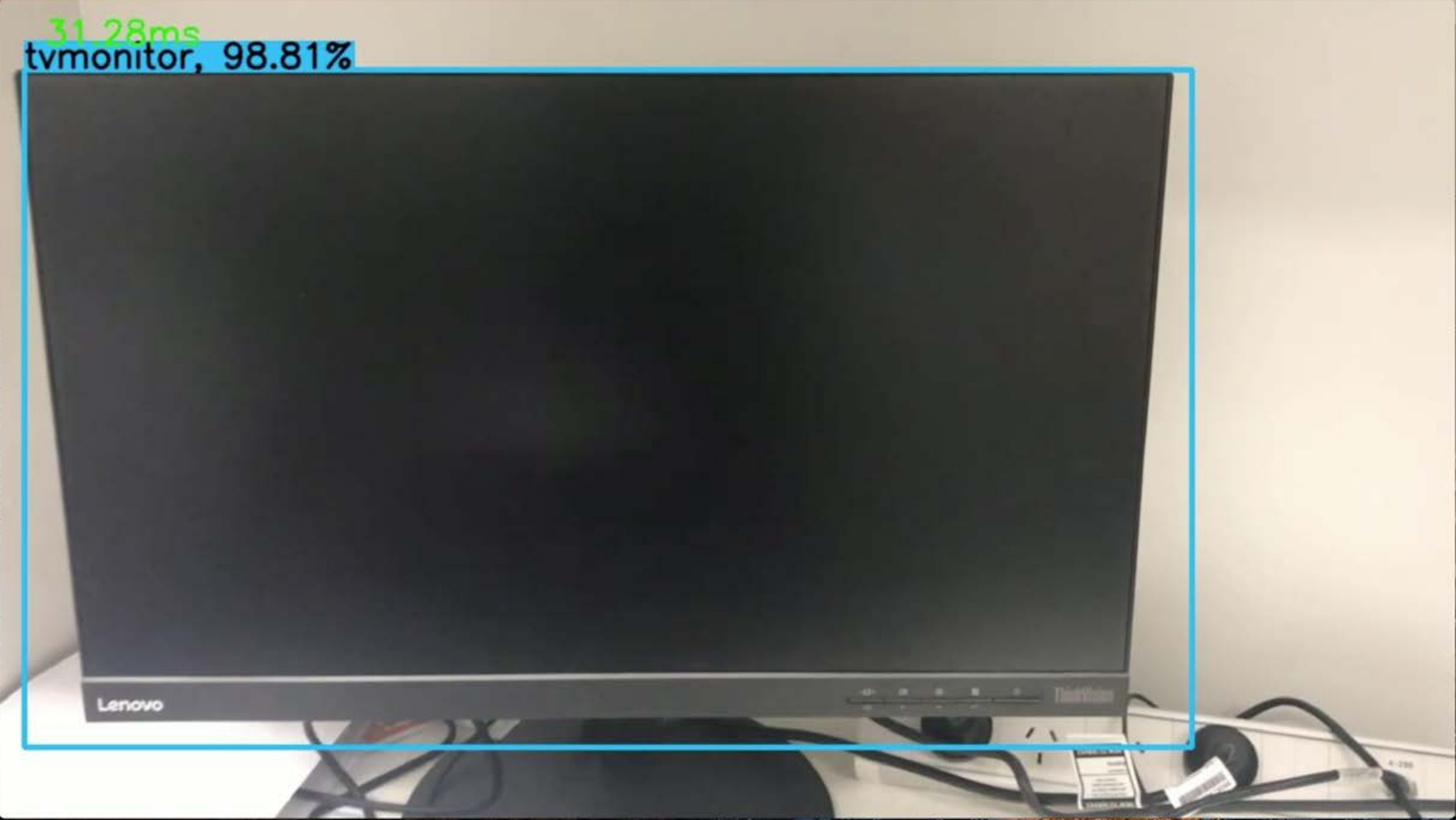}
	\caption{}
	\label{no_poster_monitor}
  \end{subfigure}%
  \begin{subfigure}[b]{.45\linewidth}
    \centering
    \includegraphics[width=1\linewidth]{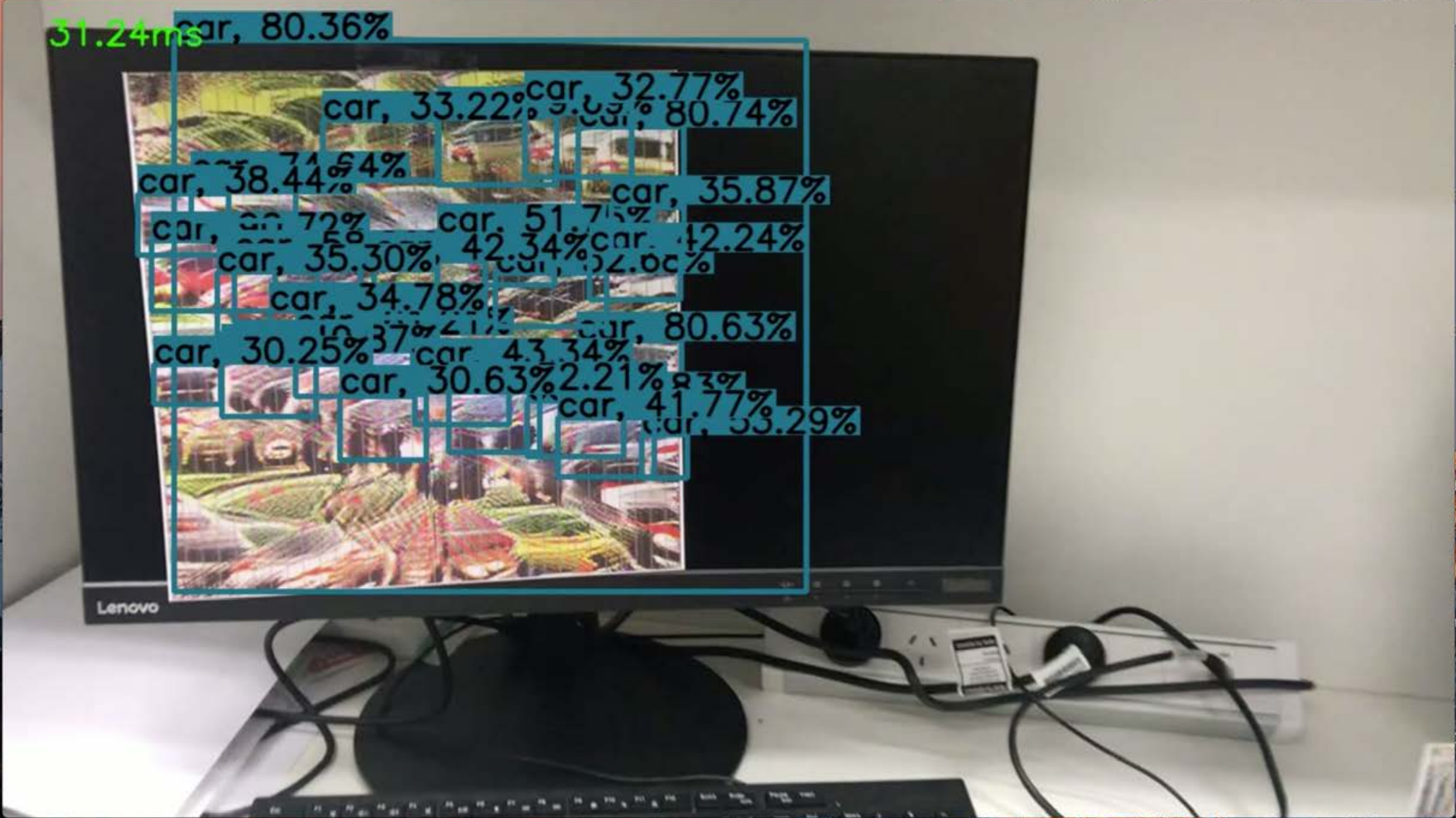}
    \caption{}
    \label{adv_monitor}
  \end{subfigure}%
  \caption{(a) Detection results of a monitor in a benign setting. (b) Detection results of the monitor when we apply a Daedalus poster attacking the `car' category.}
  \label{monitor}
\end{figure}

As shown in Fig.~\ref{mac} and Fig.~\ref{monitor}, The Daedalus attack can fool a smart inventory management (SIM) system by including false positives in the identification results\footnote{A demo video is on YouTube: https://youtu.be/U1LsTl8vufM.}.

\end{document}